\documentclass[electronic]{vgtc}             

\graphicspath{{figures/}{pictures/}{images/}{./}} 

\usepackage{times}                     

\usepackage{tabu}                      
\usepackage{booktabs}                  
\usepackage{lipsum}                    
\usepackage{mwe}                       

\usepackage{mathptmx}                  
\usepackage{graphicx}
\usepackage{booktabs}
\usepackage{multirow}
\usepackage{float}
\usepackage{makecell}
\usepackage{subcaption}
\usepackage{threeparttable}
\usepackage{wrapfig}
\usepackage{amsmath}
\usepackage{amssymb}
\usepackage{xspace}

\makeatletter
\DeclareRobustCommand\onedot{\futurelet\@let@token\@onedot}
\def\@onedot{\ifx\@let@token.\else.\null\fi\xspace}

\def\ie{\emph{i.e}\onedot}

\def\etal{\emph{et al}\onedot}
\makeatother
\onlineid{1664}

\vgtccategory{Research}

\vgtcinsertpkg

\title{CUBE360: Learning Cubic Field Representation for Monocular 360 Depth Estimation for Virtual Reality 
}

\author{Wenjie Chang \thanks{W. Chang, T. Zhang are with USTC, Hefei, China and DSEL, Hefei, China. E-mail: changwj@mail.ustc.edu.cn, tzzhang@ustc.edu.cn. This work was done when Wenjie was an intern at VLIS LAB at HKUST(GZ).} 
\and Hao Ai \thanks{H. Ai is with HKUST(GZ), Guangzhou, China. E-mail: hai033@connect.hkust-gz.edu.cn.}
\and Tianzhu Zhang \footnotemark[1]
\and Lin Wang\thanks{L. Wang is the the corresponding author and is with HKUST(GZ), Guangzhou, and HKUST, Hong Kong SAR, China. E-mail: linwang@ust.hk. \vspace{-40pt}}}
\affiliation{\textbf{\textit{Project Website:}} \url{https://wj-chang-42.github.io/cube360/}}

\teaser{
  \centering
  \includegraphics[width=0.97\linewidth]{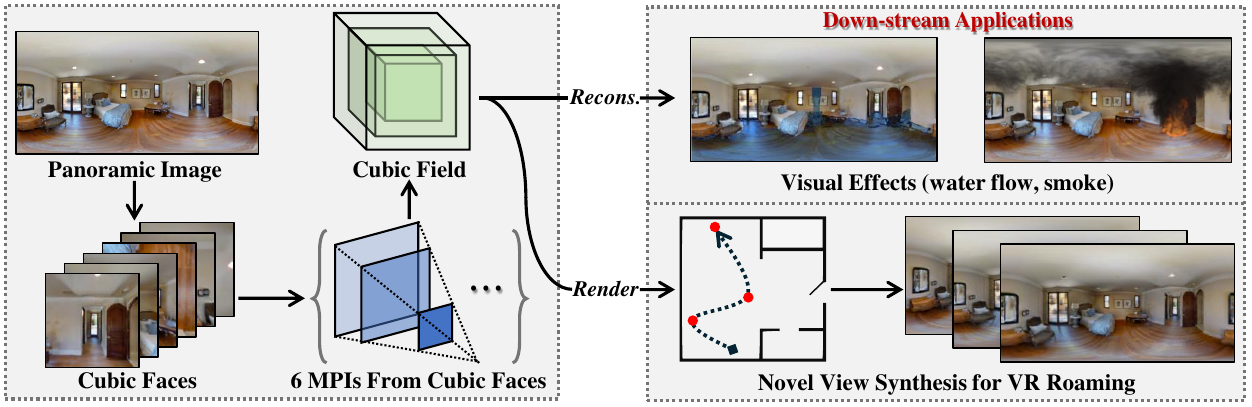}
  \caption{CUBE360 constructs a cubic field representation of a scene from one panorama captured by a 360-degree camera. 
  A single panoramic image is projected onto six cubic faces and a neural network predicts Multi-Plane Images (MPIs) for each face. These six MPIs are subsequently fused to form the cubic field, encapsulating both the density and color information of the entire scene.
  CUBE360 enables visual effects by providing depth maps that allow for realistic physical interaction within the real-world environment, while supporting VR roaming by dynamically rendering the scene from different viewpoints using ray-cube sampling and neural rendering for immersive exploration.}
  \label{fig:teaser}
}

\abstract{
    Panoramic images provide comprehensive scene information and are suitable for VR applications. Obtaining corresponding depth maps is essential for achieving immersive and interactive experiences. However, panoramic depth estimation presents significant challenges due to the severe distortion caused by equirectangular projection (ERP) and the limited availability of panoramic RGB-D datasets. 
    Inspired by the recent success of neural rendering, we propose a novel method, named \textbf{CUBE360}, that learns a cubic field composed of multiple MPIs from a single panoramic image for \textbf{continuous} depth estimation at any view direction.  
    Our CUBE360 employs cubemap projection to transform an ERP image into six faces and extract the MPIs for each, thereby reducing the memory consumption required for MPI processing of high-resolution data. Additionally, this approach avoids the computational complexity of handling the uneven pixel distribution inherent to equirectangular projection.
    An attention-based blending module is then employed to learn correlations among the MPIs of cubic faces, constructing a cubic field representation with color and density information at various depth levels. Furthermore, a novel sampling strategy is introduced for rendering novel views from the cubic field at both cubic and planar scales. The entire pipeline is trained using photometric loss calculated from rendered views within a self-supervised learning approach, enabling training on 360 videos without depth annotations. Experiments on both synthetic and real-world datasets demonstrate the superior performance of CUBE360 compared to prior SSL methods. We also highlight its effectiveness in downstream applications, such as VR roaming and visual effects, underscoring CUBE360's potential to enhance immersive experiences.

} 


\begin{document}

\firstsection{Introduction}

\maketitle


360 or panoramic cameras can capture a whole scene with a large field of view (FoV) of $180^\circ \times 360^\circ$ and are widely used in VR applications to provide immersive and interactive experiences. Since omnidirectional depth information can greatly enhance the realism and interactivity of virtual environments by accurately mapping the 3D geometry of the surrounding scene, the ability to infer depth from a single 360-degree image has driven a large suite of research endeavors for monocular 360 depth estimation. 
Existing works are predominantly supervised: they obtain the depth map directly from a single panoramic image with training on RGB-D datasets ~\cite{zioulis2018omnidepth,BiFuse20,cheng2018cube,slice,zhuang2021acdnet,SunSC21}. 

Several recent works~\cite{Wang2022BiFuseSA,zhuang2023spdet,zioulis2019spherical} have explored self-supervised panoramic depth estimation, which trains the depth estimation network by rendering images at different viewpoints and constructing photometric loss. The current self-supervised models mainly adopt image-based rendering for novel view synthesis. As depth maps fail to capture the content hidden in the reference view but revealed in the target view, rendering novel views from depth maps is insufficient, which further affects the supervision of the depth estimation network by the photometric loss.
To overcome this limitation, researchers adopt MPI representation~\cite{zhou2018stereo,tucker2020single,zhang2023structural} that models 3D space with a set of front-parallel layers to generate satisfying renderings under disocclusions and non-Lambertian effects and thus produce reasonable depth maps. To further improve MPI representation, MINE~\cite{li2021mine} generalizes MPI to a continuous 3D representation rendering schemes in Neural Radiance Field (NeRF)~\cite{mildenhall2020nerf}.

 However, the unique characteristics of panoramic images present challenges for these MPI-based 3D representations. (1) Due to the high resolution of panoramic images, generating MPI-based representations demands significant GPU memory, making the training process challenging. The method proposed in MINE~\cite{li2021mine} exemplifies this issue by outputting information for a single plane at a specific depth at a time, which requires the network to perform multiple inferences to generate MPIs at different depth levels. This not only increases computational demands but also significantly exacerbates GPU memory usage. (2) Processing panoramic images is complicated by the significant distortions introduced by equirectangular projection (ERP). Specifically, 360-degree images are displayed in 2D planar representations while preserving the omnidirectional scene details. ERP is the most common projection method for capturing a complete view of a scene but suffers from severe distortions, particularly at the poles~\cite{coors2018spherenet,yoon2022spheresr}. 
\begin{figure}[t]
    \centering
\includegraphics[width=\linewidth]{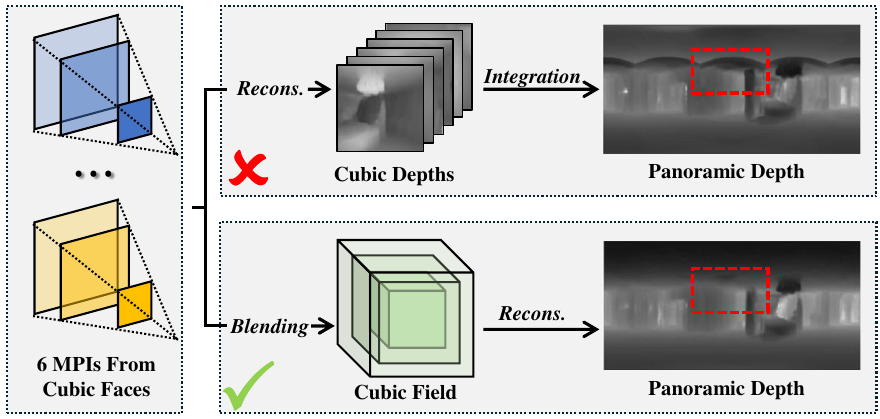}
    \captionof{figure}{We show results from MPIs and our cubic representation. The proposed cubic field produces consistent panoramic depth estimation against estimation from MPIs.}
    \label{fig:teaser_2}
    \vspace{-15pt}
\end{figure}
In contrast, cubemap projection (CP) splits the 360-degree content into six distinct 2D images, corresponding to the faces of a cube, which not only reduces distortion but also lowers the resolution of each individual image~\cite{BiFuse20, Jiang2021UniFuseUF}. Leveraging these advantages, we introduce a novel panorama representation based on cube-wise MPI, termed the cubic field. In our proposed pipeline, one panoramic image is first divided into six faces of a cubemap. An encoder-decoder based network takes these cubic faces as inputs and predicts the related MPIs that reconstruct the color and density information of a conical space at the pre-defined depths separately (Sec. \ref{sec:mpi}). Subsequently, the independent predicted MPIs of six faces are fed into a series of blending modules to generate the cubic field. These modules blend information in three ways: across different faces, between each face and the overall panorama, and along the edges where adjacent faces connect (Sec. \ref{sec:cubic}). As shown in Fig. \ref{fig:teaser_2}, these blending processes results in significantly improved depth estimation. A dual sampling strategy combined with neural rendering techniques is proposed to synthesize novel views from the cubic field at both the cubic and planar scales (Sec. \ref{sec:rendering}), which are further adopted to construct the photometric losses for supervision. We evaluate our method on synthetic and real-world datasets and show that it outperforms state-of-the-art methods in accuracy and generalization. We demonstrate that our method can produce realistic and consistent depth maps for panoramic images with various scenes and lighting conditions. 
Our contributions can be summarized into three-fold:

\begin{itemize}
    \item We propose a novel 3D representation for a single panoramic image named cubic field that models the RGB and density information of a holistic scene.
    \item We introduce a novel sampling strategy, which achieves novel view renderings at cubic and planar scales from the constructed cubic field and improves the performance of depth measurements.
    \item Experimental results demonstrate that our proposed method achieves superior performance in both quantitative and qualitative ways. Compared with SPDET \cite{zhuang2023spdet}, we achieved error reductions of 16.80\% and 24.3\% on the Matterport3D and Stanford2d3d subsets, respectively.
    \item We present the effects of the proposed cubic field in practical applications, such as visual effects and novel view synthesis for VR roaming. This demonstrates its ability to significantly enhance immersive user experiences.

\end{itemize}
\section{Related Work}
\label{sec:related}

\noindent \textbf{Panoramic Depth Estimation}
Image-based depth estimation is a fundamental problem in 3D vision \cite{Chen_2022_CVPR,bhat2021adabins,Huang:21,patil2022p3depth,ranftl2021vision,li2021revisiting}. Depth estimation from panoramas is more challenging due to the inherent spherical distortions brought by equirectangular projection. To deal with this issue, some methods propose to employ the deformable convolution filters~\cite{tateno2018distortion,Su2018KernelTN} to achieve the distortion-aware grid sampling, while some methods employ the adaptively combined dilated convolution
filters~\cite{zhuang2021acdnet} or row-wise rectangular convolution filters~\cite{zioulis2018omnidepth} to rectify the receptive field.
Recently, PanelNet~\cite{Yu2023PanelNetU3} partitions an ERP image into vertical slices and directly applies the standard convolutional layers to predict the slice-wise depth maps and then stitches them back into the ERP format. Based on the vision transformer~\cite{AlexeyDosovitskiy2020AnII}, PanoFormer~\cite{shen2022panoformer} and EGFormer~\cite{Yun2023EGformerEG} build the distortion-aware transformer blocks to process the ERP panoramas. Besides, some works~\cite{BiFuse20,Jiang2021UniFuseUF,Ai2023HRDFuseM3} introduce the bi-projection-based approaches to combine the complete view of ERP images with the local details of other less-distorted projection format input, \ie, cubemap projection (CP) and tangent projection (TP) patches. While a large body of work exists for supervised panoramic depth estimation, there exists a significant challenge for these data-driven methods, which is large-scale accurate panoramic RGB-depth pairs. Due to the large field-of-view (FoV) of the panoramas, it is expensive and challenging to collect large-scale, real-world, reliable panoramic depth datasets. As the self-supervised training strategy can get rid of the dependence on the depth ground truth, panoramic depth estimation under the self-supervised training scenario is desired. However, there exist a few methods to explore self-supervised 360 depth estimation. Inspired by~\cite{Zhou2017UnsupervisedLO}, Wang~\etal~\cite{Wang2018SelfSupervisedLO} proposed the first self-supervised framework based on the spherical photometric
consistency constraint to predict the panoramic depth maps from less-distorted CP projection patches with cubemap padding~\cite{Cheng2018CubePF}. In contrast, Zioulis~\etal~\cite{zioulis2019spherical} introduced the spherical view synthesis to estimate the depth maps. In the recently proposed BiFuse++~\cite{Wang2022BiFuseSA}, a two-stream framework, consisting of DepthNet and PoseNet, is conducted to estimate the panoramic depth based on the bi-directional feature fusion between ERP images and cubemap, and predict the camera pose from three sequential panoramas.

\noindent \textbf{Multi-Plane Image (MPI) Representation}
MPIs are a popular representation for scene reconstruction, which consists of a set of front-parallel RGBA layers that model the scene's appearance and geometry. Recent works have applied deep learning methods to generate MPIs from sparse inputs, such as stereo pairs or single images. Flynn~\etal~\cite{flynn2016deepstereo} proposed a deep-learning framework to infer MPIs from stereo pairs and synthesize novel views by blending the warped layers. Zhou~\etal~\cite{zhou2018stereo} extended this framework to handle more general camera motions and occlusion. Li~\etal~\cite{li2021mine} proposed a local light field fusion method that integrates multiple MPIs to render high-quality novel views. Zhang~\etal~\cite{zhang2023structural} introduced a transformer-based network to predict the plane poses and RGBA contexts of the Structural Multiplane Image layers and also handled non-planar regions as a particular case. However, most of these works assume that the inputs are perspective images, and thus the MPI representation is suitable for modeling the scene geometry.

A straightforward approach to adapting MPIs for panoramic scenes is the Multi-Sphere Images (MSIs), which represents the entire scene using spheres of different sizes centered at a common origin. Some recent studies have explored this representation.
MatryODShka ~\cite{attal2020matryodshka} learns MSIs from stereo setups. SOMSI~\cite{habtegebrial2022somsi} extends this concept by utilizing images from multiple viewpoints, following a similar experimental setting to NeRF. However, Several challenges arise when applying MPIs to panoramic images in a simplistic manner. Firstly, due to the sphere-to-plane projection, the ERP format panoramic images introduce significant distortion, especially near the poles. Secondly, high-resolution panoramas with the large FoV require heavy memory and computation to generate and render MPIs. Unlike previous works, our approach constructs a cubic-based representation to overcome these challenges, operating with just a single panoramic image.

\section{The Proposed CUBE360}
\begin{figure*}[t]
    \centering
    \includegraphics[width=\linewidth]{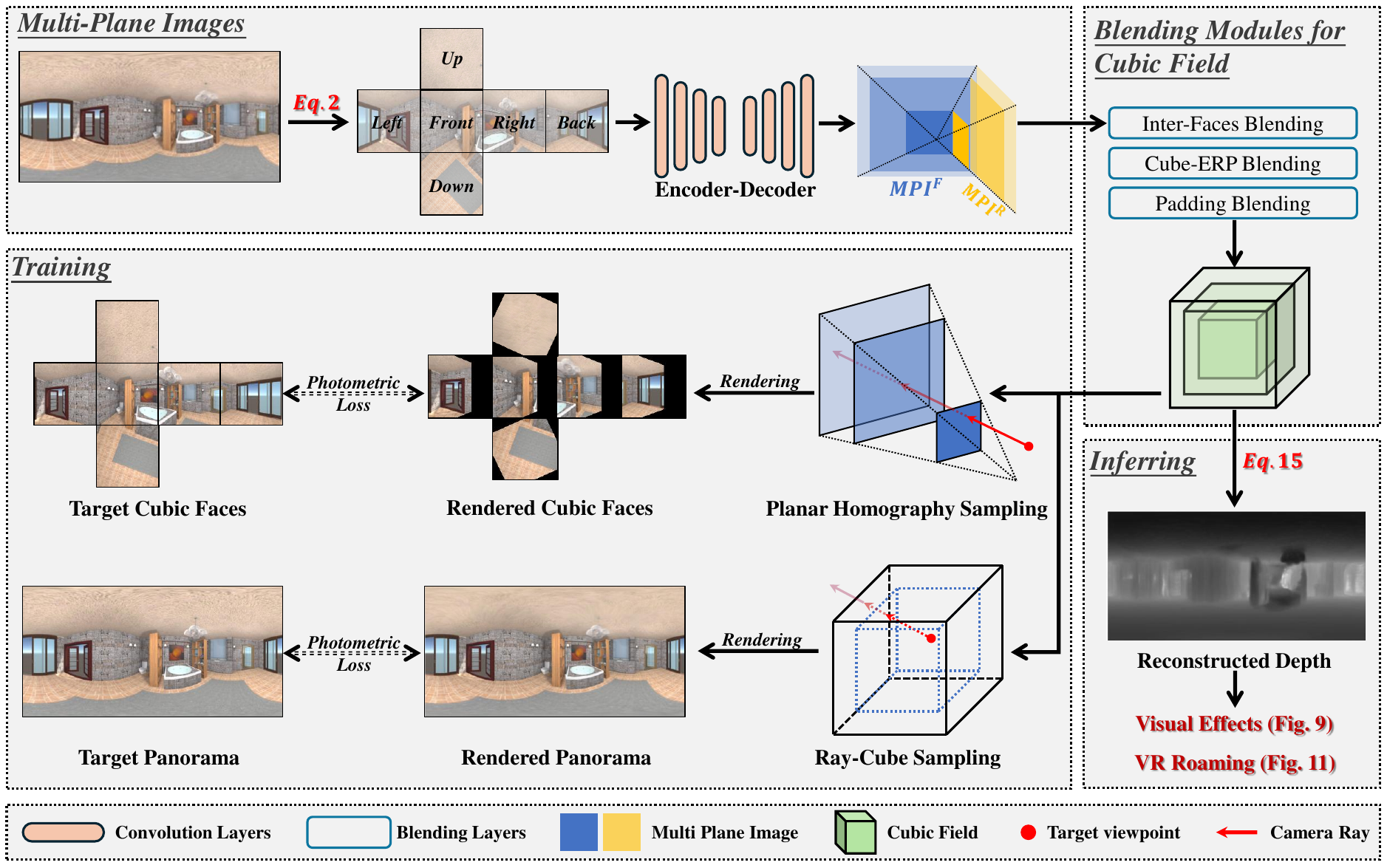}
    \caption{An overview of the proposed pipeline. An input panorama is split into six cubic faces, each capturing a different scene view. A convolutional-based network takes the cubic faces as inputs and generates the MPIs $\left\{ \mathbf{MPI}^i_0 | i \in \left\{ B, D, F, L, R, U\right\} \right\}$ for each view. The MPIs capture the scene’s appearance and geometry by representing the RGB value $c$ and density values $\sigma$ of imaging planes at a set depth levels $\mathbf{D}$. Then, a series of blending operations are proposed to update and integrate separate MPIs from different faces. The integrated features are then used to extract a cubic field, manifested as the fused MPIs $\left\{ \mathbf{MPI}^i_t | i \in \left\{ B, D, F, L, R, U\right\} \right\}$ at depth set $\mathbf{D}$.
    Novel views are rendered from the cubic field at two scales and utilized to construct photometric loss for supervision.
    }
    \label{fig:overview}
\end{figure*}

\noindent \textbf{Overview.}
The proposed CUBE360 aims to learn a novel cubic representation for the holistic scene captured by a single panorama. In this section, we offer a succinct explanation of the cubemap projection process and the representation of MPIs. These techniques are utilized to generate cubic faces and the associated MPIs. Subsequently, we present a comprehensive introduction to our proposed attention-based blending modules, illustrated in Figure~\ref{fig:overview}, which utilized to generate a cubic field from the predicted MPIs. It leverages the context and position information of MPIs to update the representation of the holistic scene.
After that, the rendering schemes for novel view synthesis are introduced, including volume rendering techniques from NeRF and our specifically designed dual sampling strategy.
The sampling strategies work on both cubic and planar scales to get cubemaps and panoramas at target viewpoints for supervision, which further facilitates the training.
Finally, we present the loss function for supervising the network to learn the cubic field. In particular, we propose a novel loss function to supervise the consistency of geometry information at cubic edges.

\subsection{Cubemap projection and Multi-Plane Images}
\label{sec:mpi}
Given a panoramic image, we first project it to the unit sphere as follows:
\begin{align}
\label{eq:e2s}
\begin{split}
&\theta = 2\pi \frac{m-c_x}{W}, \qquad \phi = \pi\frac{n-c_y}{H},\\
&\mathbf{q} = [\cos(\phi)\sin(\theta),\sin(\phi),\cos(\phi)\cos(\theta)]^T,
\end{split}
\end{align}
\noindent

where $W$ and $H$ represent the width and height of the panoramic image, respectively, $c_x$ and $c_y$ are the coordinates of the principal points. $[m,n]$ is the Image Coordinate of a pixel in the panoramic image. 
Then, the perspective projection is utilized to map the generated unit sphere to six faces of a cubemap, which is formulated as: 
\setlength\abovedisplayskip{2pt}
\setlength\belowdisplayskip{2pt}
\begin{align}
\label{eq:e2c}
\hat{\mathbf{q}} = \mathbf{K}\mathbf{r}_{i}\mathbf{q}, \quad i \in [B, D, F, L, R, U]
\end{align}
\begin{equation}
\mathbf{K}=\left[\begin{array}{ccc}
w / 2 & 0 & w / 2 \\
0 & w / 2 & w / 2 \\
0 & 0 & 1
\end{array}\right],
\end{equation}
where $K$ is the camera intrinsics of the denoted perspective projection, $w$ is the size of the cubic face, $\mathbf{r}_{i}$ is the rotation matrix to rotate a specific face to the imaging plane, and $i$ is denoted as $i$-th face, representing one of the faces of back, down, front, left, right, and up, respectively. The cubemap is obtained by repeating the above process for the six faces, which is denoted as $ \mathbf{f}_i  = E2C(\mathbf{p})$, where $\mathbf{p} \in \mathbb{R}^{H\times W}$ is the input panoramic image, $\{ \mathbf{f}_i \in \mathbb{R}^{w\times w}| i \in [B, D, F, L, R, U] \}$ are the cubic faces that are adopted for MPIs generation. 

\begin{figure}[t]
    \centering
    \includegraphics[width=\linewidth]{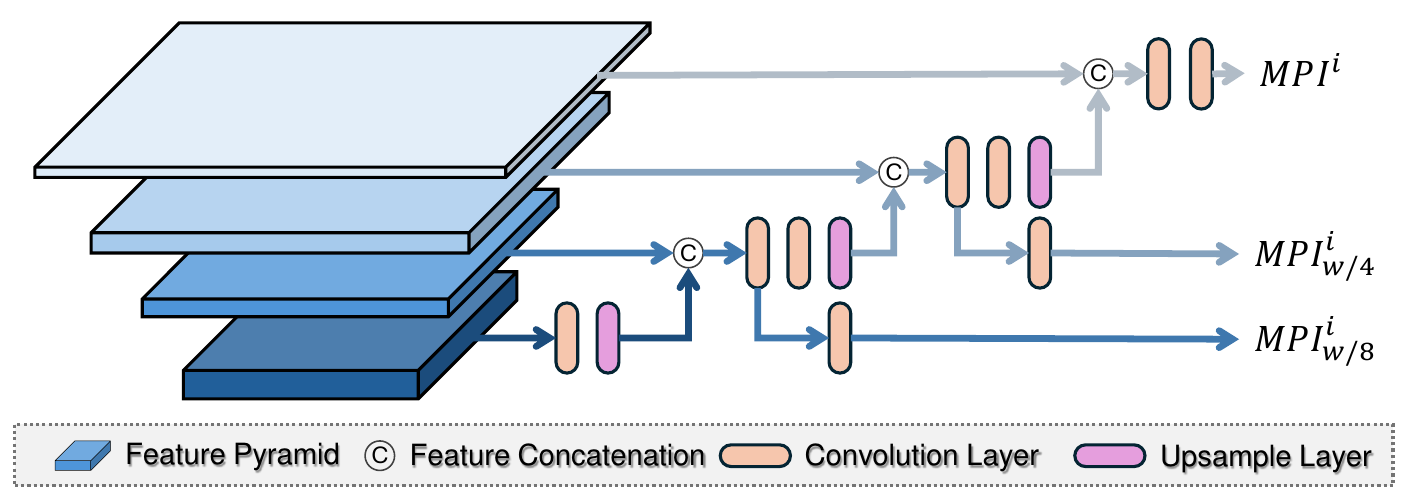}
    \captionof{figure}{\textbf{Network Details of the adopted Encoder-Decoder architecture.} $MPI_{w/4}$ and $MPI_{w/8}$ are Multi-Plane Images predicted at resolutions $[w/4,w/4]$ and $[w/8,w/8]$, respectively. $MPI_0^i$ is the predicted MPIs at the resolution $[w/2,w/2]$ and is further fed into the proposed blending modules.}
    \label{fig:en}
    \vspace{-20pt}
\end{figure}

With the cubic faces $\mathbf{f}_i$ as the inputs, MPIs are predicted by an encoder-decoder network and are denoted as
\begin{align}
    \mathbf{MPI}^i = Net(\mathbf{f}_i).
\end{align}
$\{ \mathbf{MPI}^i \in \mathbb{R}^{d\times w\times w \times 4}| i \in [B, D, F, L, R, U] \}$ is the predicted MPIs for each face. 
An $\mathbf{MPI}^i$ includes the radiance $c_{z} \in \mathbb{R}^{w\times w \times 3}$ and density $\sigma_{z} \in \mathbb{R}^{w\times w \times 1}$ of $d$ planes, where $z$ denotes the pre-defined depth value of the related plane. We denote the set of $d$ depth values as $D=\{z_1,z_2,\dots , z_d\}$. Specifically, we adopt an encoder-decoder architecture for our model. The encoder is a ResNet-50 network that produces a feature pyramid from its intermediate layers. The decoder consists of convolutional and upsampling layers that generate multiplane images (MPIs) at different scales, as illustrated in Figure~\ref{fig:en}. These MPIs are then used to synthesize novel views at various resolutions. In addition, the output of the topmost layer in the decoder is fed into the proposed blending modules.

\subsection{Cubic Field Representation.}

\begin{figure*}[t]
    \centering
    \includegraphics[width=\linewidth]{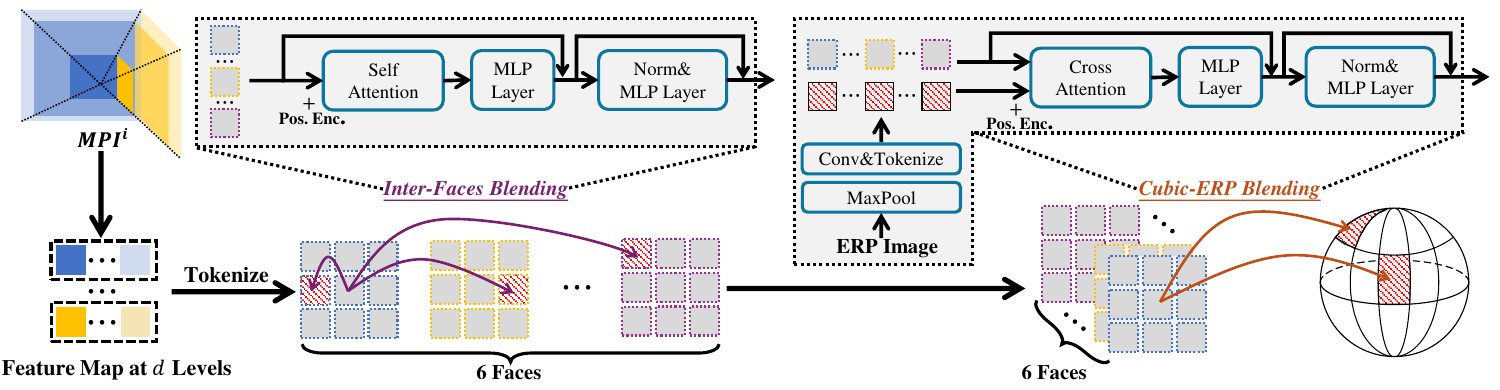}
    \caption{Illustration of the inter-face blending and cube-ERP blending processes in CUBE360. For inter-face blending, the Multi-Plane Images (MPIs) from the six cubic faces are tokenized and fed into a self-attention module to enhance the holistic representation of the cubic field. The positional encoding is applied based on spherical coordinates, and the resulting attention matrix helps capture interactions between tokens. In the cube-ERP blending stage, global ERP features, extracted through convolution and pooling, are integrated with the cubic field tokens using cross-attention. The final output restores six feature maps, representing the enhanced geometry and color information for each cubic face.
    }
    \label{fig:blending}
\end{figure*}

\begin{figure}[t]
    \centering
    \begin{subfigure}{\linewidth}
    \centering
    \includegraphics[width=\linewidth]{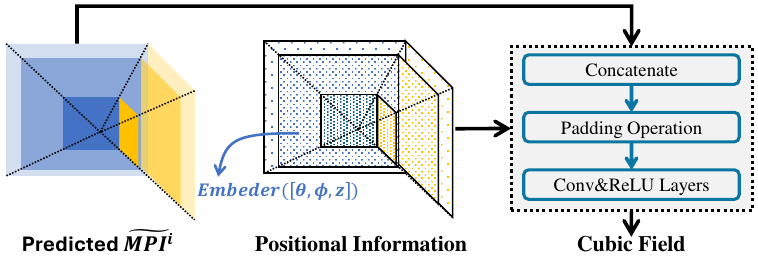}
    \caption{Padding Blending}
    \label{fig:padding:a} 
    \end{subfigure}
    \vfill
    \begin{subfigure}{0.8\linewidth}
    \centering
    \includegraphics[width=\linewidth]{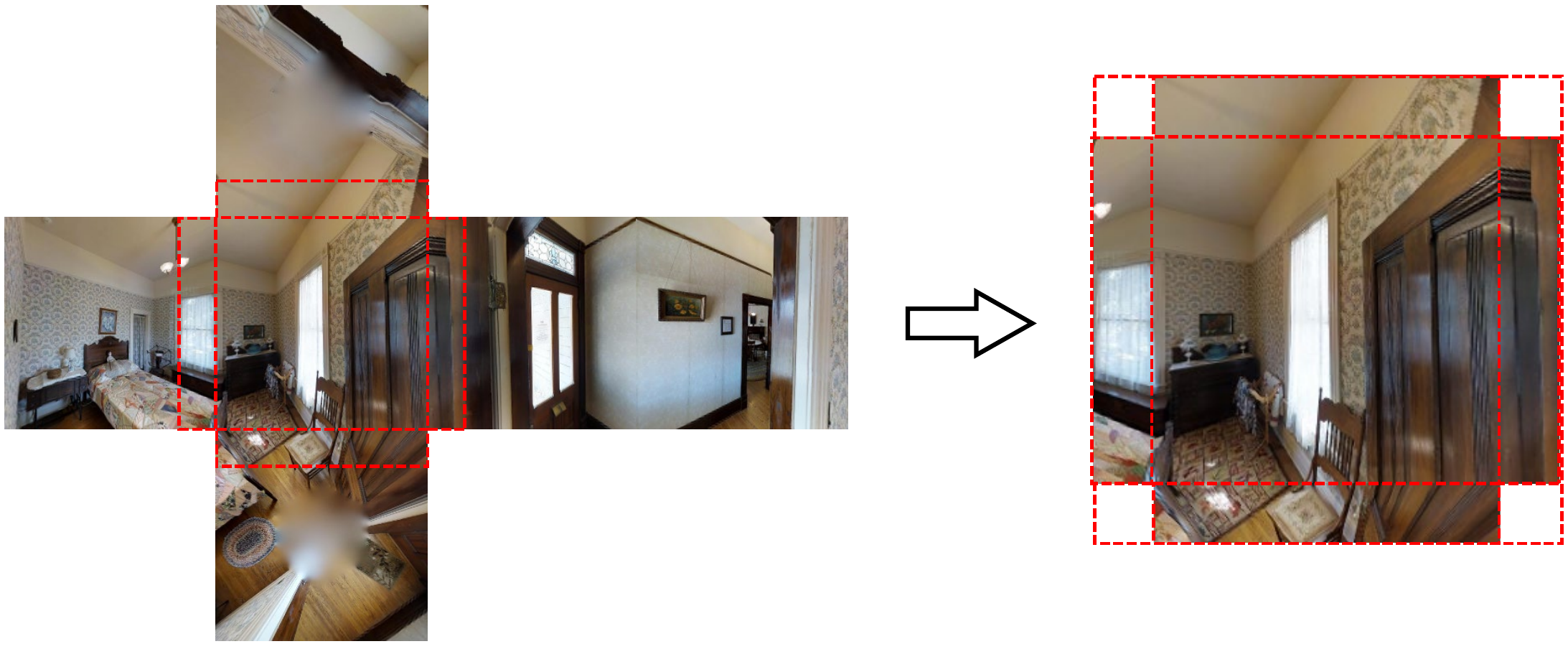}
    \caption{Padding Operation}
    \label{fig:padding:b} 
    \end{subfigure}
    \caption{Illustration of the padding blending. \textcolor{red}{(a)} represents the proposed padding blending method, where $\tilde{\mathbf{MPI}}^i$ is concatenated with the corresponding positional information, followed by convolution and ReLU activation to generate the cubic field.
    In \textcolor{red}{(b)}, the left image illustrates the adjacency relationship between the target cubic face and the other five faces. As depicted in the right image, this adjacency relationship enables us to integrate the information at the edge of the adjacent face into the target cubic face, thereby achieving feature fusion at the cubic level.}
    \label{fig:padding}
    \vspace{-23pt}
\end{figure}

\label{sec:cubic}
\subsubsection{Inter Face Blending.}
\label{sec:face-blending}
To integrate information across the different faces of the cubic representation, we employ an inter-face blending module as shown in Fig. \ref{fig:blending}. Given the set of Multi-Plane Images ${ \mathbf{MPI}^i \in \mathbb{R}^{d\times w\times w \times 4} \mid i \in [B, D, F, L, R, U] }$, each image plane is divided into 16x16 patches, which are then flattened into tokens with dimensions $d \times \frac{w}{16} \times \frac{w}{16} \times (4 \cdot 16 \cdot 16)$. These tokens capture the features extracted from each image plane. Subsequently, the tokens from the six different faces of the cube are integrated as $\mathbf{z} \in  \mathbb{R}^{d \times \frac{6w^2}{256} \times 1024}$ that are subsequently fed into the inter-faces blending module. In particular, the self-attention mechanism ($SA$) is utilized to calculate calculate the interactions between these tokens to enhance the holistic representation of the cubic field. In the self-attention module, the input sequence $\mathbf{z}$, combined with positional encoding $pos{c}$, is projected through three different weight matrices as follows: 
\begin{align}
\mathbf{q} = (\mathbf{z}+ pos_{c}) W_q,\;
\mathbf{k} = (\mathbf{z} + pos_{c}) W_k,\;
\mathbf{v} = \mathbf{z} W_v,
\end{align}
where $\mathbf{q}, \mathbf{k}, \mathbf{v} \in \mathbb{R}^{N \times M}$ represent the query, key, and value embeddings, respectively. Positional embedding $pos_{c}$ provides information about the position of each token within the sequence. For each token, its center coordinates $[\theta, \phi]$ on the unit sphere is derived by applying Eq.~\ref{eq:e2s} and \ref{eq:e2c}. The embedding vector for each token is then computed using sinusoidal functions based on these coordinates. Specifically, for each index $i=0,1,...,255$, the embedding vector components are defined as follows:
\begin{align}
\begin{aligned}
\left[ 
\cos\left(\frac{\theta \cdot \pi}{10000^{\frac{i}{256}}}\right), 
\sin\left(\frac{\theta \cdot \pi}{10000^{\frac{i}{256}}}\right), \right. \\
\left. \cos\left(\frac{\phi \cdot \pi/2}{10000^{\frac{i}{256}}}\right), 
\sin\left(\frac{\phi \cdot \pi/2}{10000^{\frac{i}{256}}}\right) 
\right]
\end{aligned}
\end{align}
Then, the attention matrix $\mathbf{A}$, which captures the similarity between tokens at different positions, is computed as:
\begin{equation}
    \mathbf{A} = \text{SOFTMAX}\left(\frac{\mathbf{q}\mathbf{k}^\top}{\sqrt{M}}\right).
\end{equation}
The output of the self-attention mechanism is an aggregation of the values weighted by the attention scores:
\begin{equation}
    SA(\mathbf{z}) = \mathbf{A} \mathbf{v}.
\end{equation}
Then, the output of the self-attention mechanism $SA(\mathbf{z})$ passes through two fully connected layers followed by a skip connection to produce the output of the module, denoted as $\hat{\mathbf{z}}$.

\subsubsection{Cube-ERP Blending}
The predicted $\hat{\mathbf{z}}$ are further integrated with the global ERP information to enhance the overall representation. As illustrated in Fig. \ref{fig:blending}, the ERP is first processed through convolution and pooling operations to reduce its dimensionality and resolution and then divided into tokens, producing $\mathbf{z}_{erp} \in \mathbb{R}^{1\times H/32\cdot W/32 \times 1024}$ denoted as $\mathbf{z}_{erp}$. Subsequently, the Cross-Attention mechanism ($CA$) integrates the global ERP features with the tokens $\hat{\mathbf{z}}$. The specific operations are as follows:
\begin{align}
\mathbf{q} = (\mathbf{\hat{z}}+ pos_{c}) W_q,\;
\mathbf{k} = (\mathbf{z}_{erp} + pos_{e}) W_k,\;
\mathbf{v} = \mathbf{z}_{erp} W_v,
\end{align}
where $\mathbf{q}, \mathbf{k}, \mathbf{v} \in \mathbb{R}^{N \times M}$ represent the query, key, and value embeddings, respectively. The calculation of $pos_{e}$ is identical to that of $pos_{c}$ described in Sec. \ref{sec:face-blending}.
The attention matrix $\mathbf{A}$, which captures the similarity between $\mathbf{\hat{z}}$ and $\mathbf{z}_{erp}$, is computed as:
\begin{equation}
    \mathbf{A} = \text{SOFTMAX}\left(\frac{\mathbf{q}\mathbf{k}^\top}{\sqrt{M}}\right), \quad \mathbf{A} \in \mathbb{R}^{N \times N}.
\end{equation}
The output of the self-attention mechanism is an aggregation of the values weighted by the attention scores:
\begin{equation}
    CA(\hat{\mathbf{z}},\mathbf{z}_{erp}) = \mathbf{A} \mathbf{v}.
\end{equation}
Then, the output of the cross-attention mechanism $CA(\hat{\mathbf{z}},\mathbf{z}_{erp})$ passes through two fully connected layers followed by a skip connection to produce the output of the module, denoted as $\tilde{\mathbf{z}}\in \mathbb{R}^{d\times \frac{6w^2}{256} \times 1024}$. Finally, $\tilde{\mathbf{z}}$ is reverse split to restore the six feature maps, resulting in $\{ \tilde{\mathbf{MPI}}^i \in \mathbb{R}^{d\times w\times w \times 4} \mid i \in [B, D, F, L, R, U] \}$.

\subsubsection{Padding Blending}
Given a point $\hat{\mathbf{q}}$ on the predicted $\tilde{\mathbf{MPI}^i}$, we can derive its coordinates $[\theta,\phi]$ in the unit sphere coordinate by applying Eq.~\ref{eq:e2s},\ref{eq:e2c}. Hence, we formulate a representation for a point on $\tilde{\mathbf{MPI}^i}$ by incorporating the spatial relationships of the different planes as $[c,\sigma,\theta,\phi,1/z,\gamma([\theta,\phi,1/z])]$, where $z$ is the depth of the related image plane and $\gamma$ is the positional encoding from NeRF~\cite{wei2021nerfingmvs,rahaman2019spectral}.
\begin{align}
    \gamma(\mathbf{u}) = \left[\cos(2\pi \mathbf{u}), \sin(2\pi \mathbf{u})\right],
\end{align}
where $\mathbf{u} = [\theta,\phi,1/z]$. To enhance the embedded MPI, we adopt the cube padding operation~\cite{cheng2018cube}, which pads the edges of each face with adjacent regions from neighboring faces, as illustrated in Fig. ~\ref{fig:padding}. 
This operation ensures geometric continuity across the cubemap faces in 3D space.
Then, the padding MPI is processed through convolutional layers, resulting in the final cubic field representation.

\subsection{Novel View Rendering}
\label{sec:rendering}
In this section, we present the schemes for obtaining novel view rendering that is further adopted in the construction of photometric loss. We first review the schemes of neural rendering from the proposed cubic field. Next, we introduce sampling strategies at the planar (Planar Homography Sampling) and cubic (Ray-Cube Sampling) levels respectively, which obtain the RGB and density information required for neural rendering at different viewpoints.
\begin{figure*}[t]
    \raggedright
    \begin{subfigure}{0.56\linewidth}
    \centering
    \includegraphics[height=1.35in]{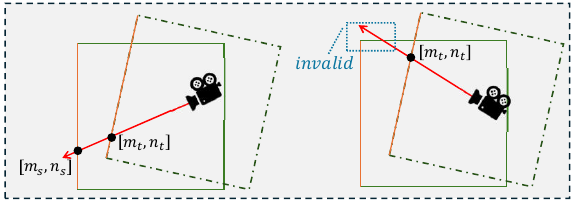}
    \caption{Planar Homography Sampling}
    \label{fig:sample:a} 
    \end{subfigure}
    \hspace{-0.18in}
    \begin{subfigure}{0.28\linewidth}
    \centering
    \includegraphics[height=1.35in]{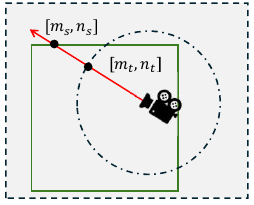}
    \caption{Ray-Cube Sampling}
    \label{fig:sample:b}
    \end{subfigure}
    \hspace{-0.18in}
    \begin{subfigure}{0.18\linewidth}
    \centering
    \includegraphics[height=1.35in]{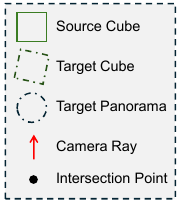}
    \vspace{15pt}
    \label{fig:sample:c}
    \end{subfigure}
    \caption{Illustration of the two sampling strategies in 2D. In \textcolor{red}{(a)}, we establish a mapping relationship between the point on the source cubic plane $[m_s,n_s]$ and a point on the target cubic plane $[m_t,m_t]$. However, when the difference between the two viewpoints is large, the mapping relationship between the plane cannot be established well. In \textcolor{red}{(b)}, we directly build the mapping of the point on the source cubic plane $[m_s,n_s]$ and the point on a unit sphere at the target view.}
\end{figure*}

{\noindent \textbf{Volume Rendering.}}
First, we illustrate the rendering mechanism for the cubic field representation. The RGB image is rendered based on the principle of classical volume rendering~\cite{mildenhall2020nerf}:
\begin{align}
\label{eq:rendering}
\hat{\mathrm{I}}=\sum_{b=1}^d T_b\left(1-\exp \left(-\sigma_{z_b} \delta_{z_b}\right)\right) c_{z_b},
\end{align}
where $T_b=\exp \left(-\sum_{j=1}^{b-1} \sigma_{z_j} \delta_{z_j}\right)$ is the map of accumulated transmittance from the closest plane to plane at depth $z_b$. Specifically, $T_b(m, n)$ denotes the probability of a ray traveling from $\left(m, n, z_1\right)$ to $\left(m, n, z_b\right)$ without hitting any object. Furthermore, the distance map between plane $b+1$ and $b$ is
\begin{equation}
\footnotesize
\setlength\abovedisplayskip{2.5pt}
\setlength\belowdisplayskip{2.5pt}
    \delta_{z_i}(m, n)=\left\|\mathcal{T}\left(\left[m, n, z_{b+1}\right]^{\top}\right) -\mathcal{T}\left(\left[m, n, z_b\right]^{\top}\right)\right\|_2
\end{equation}
where $\mathcal{T}$ represents the transformation from image coordinates to camera coordinates. Meanwhile, the depth map is extracted from the MPIs with the following:
\begin{align}
\label{eq:depth}
\hat{\mathrm{D}}=\sum_{b=1}^d T_b\left(1-\exp \left(-\sigma_{z_b} \delta_{z_b}\right)\right) F_{z_b},
\end{align}
where $F_{z_b}(m,n) = \left\|\mathcal{T}\left(\left[m, n, z_{b}\right]^{\top}\right)\right\|_2$ is the distance between the pixel $[m,n]$ of plane at depth $z_b$ and the camera origin.

{\noindent \textbf{Planar Homography Sampling.}}
We combine the standard inverse homography and the generation of cubemap images to define the Planar Homography Sampling~\cite{14,tucker2020single}. As shown in Figure \ref{fig:sample:a}, This sampling is adopted to find the correspondence between a pixel coordinate in the $i^{th}$ target cubic face $[m_t,n_t]$ and a pixel coordinate $[m_s,n_s]$ in the $i^{th}$ source cubic face and is formulated as 
\begin{equation}
\label{eq:cube}
\setlength\abovedisplayskip{3pt}
\setlength\belowdisplayskip{3pt}
    \left[m_t, n_t, 1\right]^{\top} = \mathbf{K} \left(\mathbf{r}_i\mathrm{R}\mathbf{r}_i^{\top}-\frac{\mathbf{r}_i\mathbf{t n}^{\top}}{{z}} \right) \mathbf{K}^{-1}\left[m_s, n_s, 1\right]^{\top},
\end{equation}
where $n = [0, 0, 1]^{\top}$ is the normal vector of the front parallel plane and $z$ is depth . $\mathbf{r}_i$ denotes the rotation matrix and $\mathbf{K}$ represents the camera intrinsics, which are both introduced in Eq.~\ref{eq:e2c} for generating the $i^{th}$ cubic. $\mathrm{R}$ is the rotation matrix and $\mathbf{t}$ is the translation vector. Inverting Eq.~\ref{eq:cube}, we could build the mapping $c_{z}(m_t, n_t) = c_{z}(m_s, n_s)$  and $\sigma_{z}(m_t, n_t) = \sigma_{z}(m_t, n_t)$, with which the synthesized target cubic face $\hat{\mathbf{f}}$ are rendered from the introduced volume rendering scheme.

{\noindent \textbf{Ray-Cube Sampling.}}
With the introduced Planar Homography Sampling, we build up the mapping of pixels of each cubic face at different viewpoints. However, when the camera motion is large, some pixels after Homography Sampling do not have valid values (right image of Figure \ref{fig:sample:a}), resulting in the black area of rendered cubic faces shown in Figure \ref{fig:overview}. To overcome this limitation, we utilize Ray-Cube Sampling to create a holistic representation of the target view directly from the learned cubic representation. For a pixel coordinate $[m_t,n_t]^{\top}$ at the target view, the camera ray passing through this point is represented by $\mathbf{C}_t(\rho) = \rho\mathbf{q}$, where $\mathbf{q}$ is the point on a unit sphere introduced in Eq. \ref{eq:e2s}. Then the denoted camera ray is transformed to the source view with rotation matrix $\mathrm{R}$ and translation $\mathbf{t}$ and is formulated as:
\begin{align}
    \mathbf{C}_r(\rho) = -\mathbf{t} + \rho\mathrm{R}\mathbf{q}
\end{align}
Then, the intersection of the camera ray $\mathbf{C}_r(\rho)$ and a cube of size $z$ in the source view is calculated as:
\begin{align}
\rho_{+}^{a} = \frac{{z}+ \mathbf{t}^a}{(\mathrm{R}\mathbf{q})^a},
\rho_{-}^{a} = \frac{-{z} +\mathbf{t}^a}{(\mathrm{R}\mathbf{q})^a},a={0,1,2}
\end{align}
where $a$ denotes the value of the $a^{th}$ element in the vector.
The intersection is $\mathbf{C}_r(\rho_{min})$ and the $\rho_{min}$ is the minimum positive value in $\{\ \rho_{+}^{a},\rho_{-}^{a} | a = 0,1,2 \}$. And the pixel coordinate $[m_s, n_s]$ hit by the camera ray $\mathbf{C}_r(\rho)$ at source view is calculated as:
\begin{equation}
\setlength\abovedisplayskip{3pt}
\setlength\belowdisplayskip{3pt}
\begin{aligned}
\label{eq:1314}
    [m_s,n_s]^{\top} = [W\frac{\theta}{2\pi}+c_x,H\frac{\phi}{\pi}+c_y]^{\top} 
    \\
    \phi = \sin^{-1}(\frac{{\mathbf{C}_r(\rho_{min})}^y}{|{\mathbf{C}_r(\rho_{min})}|}),\theta = \tan^{-1}(\frac{{\mathbf{C}_r(\rho_{min})}^x}{{\mathbf{C}_r(\rho_{min})}^{z}}),
\end{aligned}
\end{equation}
where $W$ and $H$ denote the width and height of the target panoramic image. Then, the synthesized target panoramic image $\hat{\mathbf{I}}_{tgt}$ is rendered from the introduced volume rendering scheme with the correspondence, $c_{z}(m_t, n_t) = c_{z}(m_s, n_s)$ and $\sigma_{z}(m_t, n_t) = \sigma_{z}(m_t, n_t)$.

\subsection{Loss Function}
\label{sec:loss}
{\noindent \textbf{Photometric loss.}}
The photometric loss minimizes the difference between the target image and the rendered image, which is formulated as:
\begin{align}
\mathcal{L}_{\mathrm{L} 1}=\frac{1}{3 H W} \sum\left|\hat{\mathbf{I}}-\mathbf{I}\right| + \frac{1}{6 w^2} \sum\left|\hat{\mathbf{f}}-E2C(\mathbf{I})\right|\\
\mathcal{L}_{\text {ssim }}=1-\operatorname{SSIM}\left(\hat{\mathbf{I}}, \mathbf{I}\right)+ 1-\operatorname{SSIM}\left(\hat{\mathbf{f}}, E2C(\mathbf{I})\right),
\end{align}
where $\mathbf{I}$ is the target panorama, $E2C(\mathbf{I})$ are target cubic faces, $\hat{\mathbf{I}}$ is the rendered panorama and $\hat{\mathbf{f}}$ are rendered cubic faces.

\begin{table*}[!t]
\begin{center}
\caption{\textbf{Evaluation results on PanoSUNCG dataset, Matterport3D and Stanford2D3D}. We evaluated the depth measurement results for the PanoSUNCG dataset between 1m and 10m. Meanwhile, for Matterport3D and Stanford2D3D, we evaluated the distance between 0.3 and 10m. \textbf{Bold} represents the best result, and \colorbox{lightgray}{Gray} indicates the second-best result.
}

\label{table:panosuncg}
\resizebox{\textwidth}{!}{ 
\setlength{\tabcolsep}{4mm}{
\begin{tabular}{cccccccc}
\hline\noalign{\smallskip}
Method & Dataset&MAE $\downarrow$ &MRE $\downarrow$ & RMSE $\downarrow$ & $\delta_1 \uparrow$& $\delta_2 \uparrow$& $\delta_3 \uparrow$ \\
\noalign{\smallskip}
\hline
\noalign{\smallskip} 
MINE&\multirow{5}{*}{\makecell[c]{PanoSUNCG\\$0.3m \sim 10m$}}&0.1876&0.1122&0.4190&0.8703&0.9529&0.9788 \\
BiFuse++&&0.1918&\textbf{0.0981}& 0.4496&\colorbox{lightgray}{0.8885}&0.9514&0.9738
 \\ 
ours w/o Blending&&\colorbox{lightgray}{0.1819}&0.1107&\colorbox{lightgray}{0.4150}&0.8777&\colorbox{lightgray}{0.9552}&\colorbox{lightgray}{0.9798}\\
ours&&\textbf{0.1673}&\colorbox{lightgray}{0.1015}&\textbf{0.3839}&\textbf{0.8970}&\textbf{0.9625}&\textbf{0.9832}
\\
\noalign{\smallskip} 
\hline
\noalign{\smallskip} 
MINE&\multirow{5}{*}{\makecell[c]{Matterport3D\\$0.3m \sim 10m$}}&0.2196&0.0980&0.4079&0.9121&\colorbox{lightgray}{0.9823}&\colorbox{lightgray}{0.9942} \\
BiFuse++&&0.2684&0.1141&0.5173&0.8672&0.9580&0.9798
 \\ 
SPDET&&\colorbox{lightgray}{0.1913}&\textbf{0.0798}&0.4028&\textbf{0.9256}&0.9790&0.9916
 \\
ours w/o Blending&&0.1995 &0.0932 &\colorbox{lightgray}{0.3678} &0.9133 &0.9832 &0.9948\\
ours&&\textbf{0.1828}&\colorbox{lightgray}{0.0867}&\textbf{0.3349}&\colorbox{lightgray}{0.9248}&\textbf{0.9855}&\textbf{0.9955}
\\

\noalign{\smallskip} 
\hline 
\noalign{\smallskip} 
MINE&\multirow{5}{*}{\makecell[c]{Stanford2D3D\\$0.3m \sim 10m$}}&0.2371&0.1046&0.4218&0.9092&\colorbox{lightgray}{0.9799}&0.9928 \\
BiFuse++&&0.2912&0.1339&0.5134&0.8672&0.9580&0.9798
 \\ 
SPDET&&0.2204&\colorbox{lightgray}{0.0940}&0.4240&0.9026&0.9739&0.9898
 \\
ours w/o Blending&&\colorbox{lightgray}{0.1992} &0.0959&\colorbox{lightgray}{0.3568} &\colorbox{lightgray}{0.9094} &0.9784 &\colorbox{lightgray}{0.9932}\\
ours&&\textbf{0.1845}&\textbf{0.0920}&\textbf{0.3209}&\textbf{0.9188}&\textbf{0.9801}&\textbf{0.9940}
\\
\noalign{\smallskip} 
\hline
\end{tabular}}}
\vspace{-10pt}
\end{center}
\end{table*}



{\noindent \textbf{Edge alignment loss.}}
In the process of transforming a panoramic image into a cubemap representation, we assume that each face of the cubemap is an independent planar projection. However, this assumption neglects the fact that the cubemap is a 3D object with continuous geometric information across the edges of adjacent faces. To address this issue, we introduce a novel loss function that penalizes the inconsistency of geometric information along the edges of the cubemap. We first extract the edge of each cubic face and calculate their weights for rendering, expressed as 
\begin{align}
    E=[w_1,\dots w_i,\dots w_d],E \in \mathbb{R}^{w\times d},\\
    w_b = T_b\left(1-\exp \left(-\sigma_{z_b} \delta_{z_b}\right)\right).
\end{align}
Then the edge align loss is 
\begin{align}
\mathcal{L}_e = cos(E,\hat{E}) + MAE(E,\hat{E}), 
\end{align}
where $cos$ denotes the calculation of cosine similarity and $MAE$ represents the Mean Absolute Error. $E$ and $\hat{E}$ are two adjacency edges in 3D space of a cube.

Then, the total loss is given by:
\begin{align}
\mathcal{L}=\lambda_{\mathrm{L} 1} \mathcal{L}_{\mathrm{L} 1}+\lambda_{\text {ssim}} \mathcal{L}_{\text {ssim}}+\lambda_e \mathcal{L}_e,
\end{align}
where $\lambda_{\mathrm{L}1}, \lambda_{\text {ssim}}, $ and $\lambda_e$ are factors of different loss terms.

\newcommand{\piclength}{0.185\textwidth}
\begin{figure*}[!t]
    \raisebox{0.35in}{\rotatebox{90}{{\scriptsize PanoSUNCG}}}
    \centering
    \begin{subfigure}{\piclength}
    \label{fig:suncg:rgb} 
    \includegraphics[width=\textwidth]{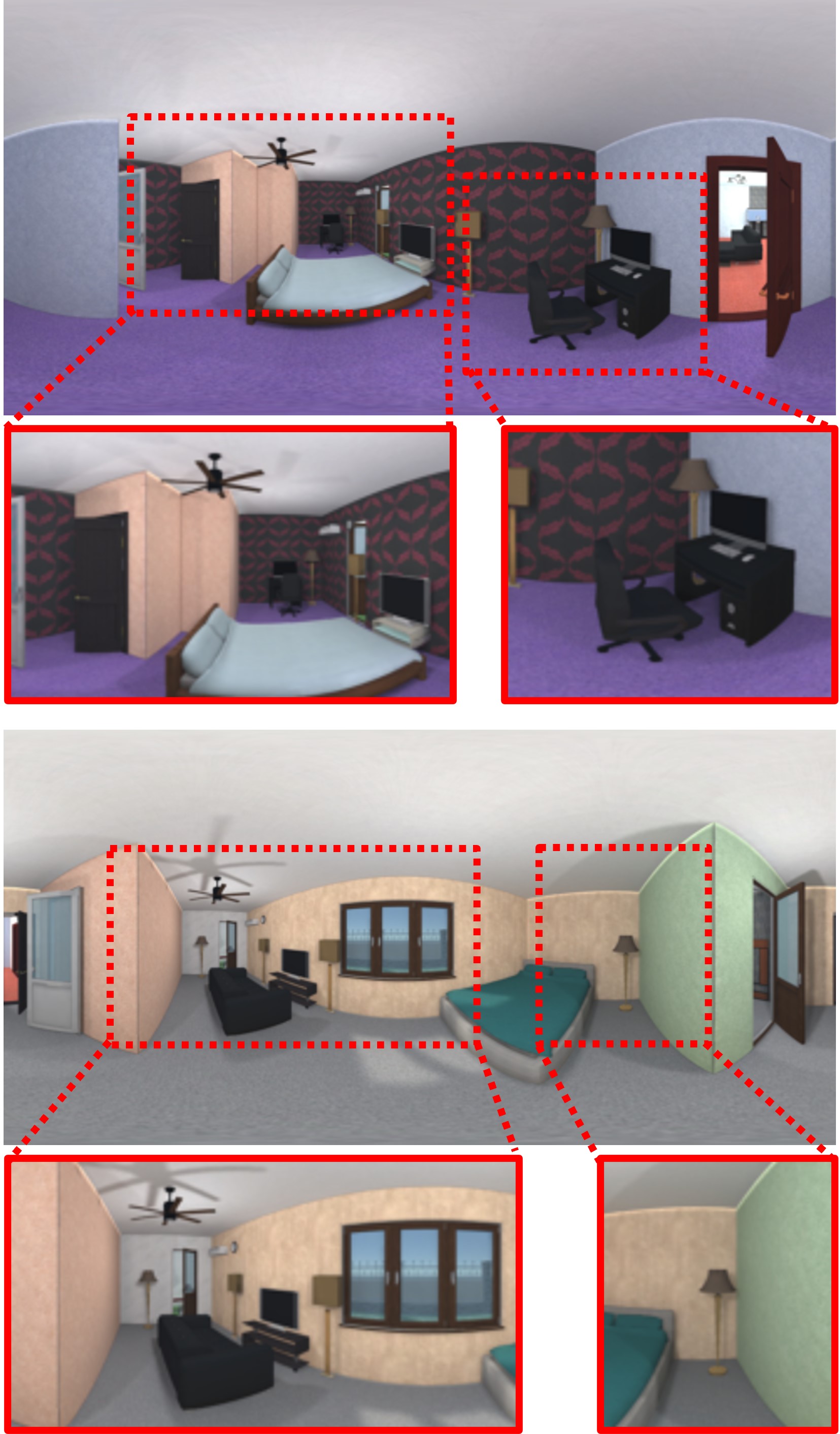}
    \end{subfigure}
    \begin{subfigure}{\piclength}
    \label{fig:suncg:gt}
    \includegraphics[width=\textwidth]{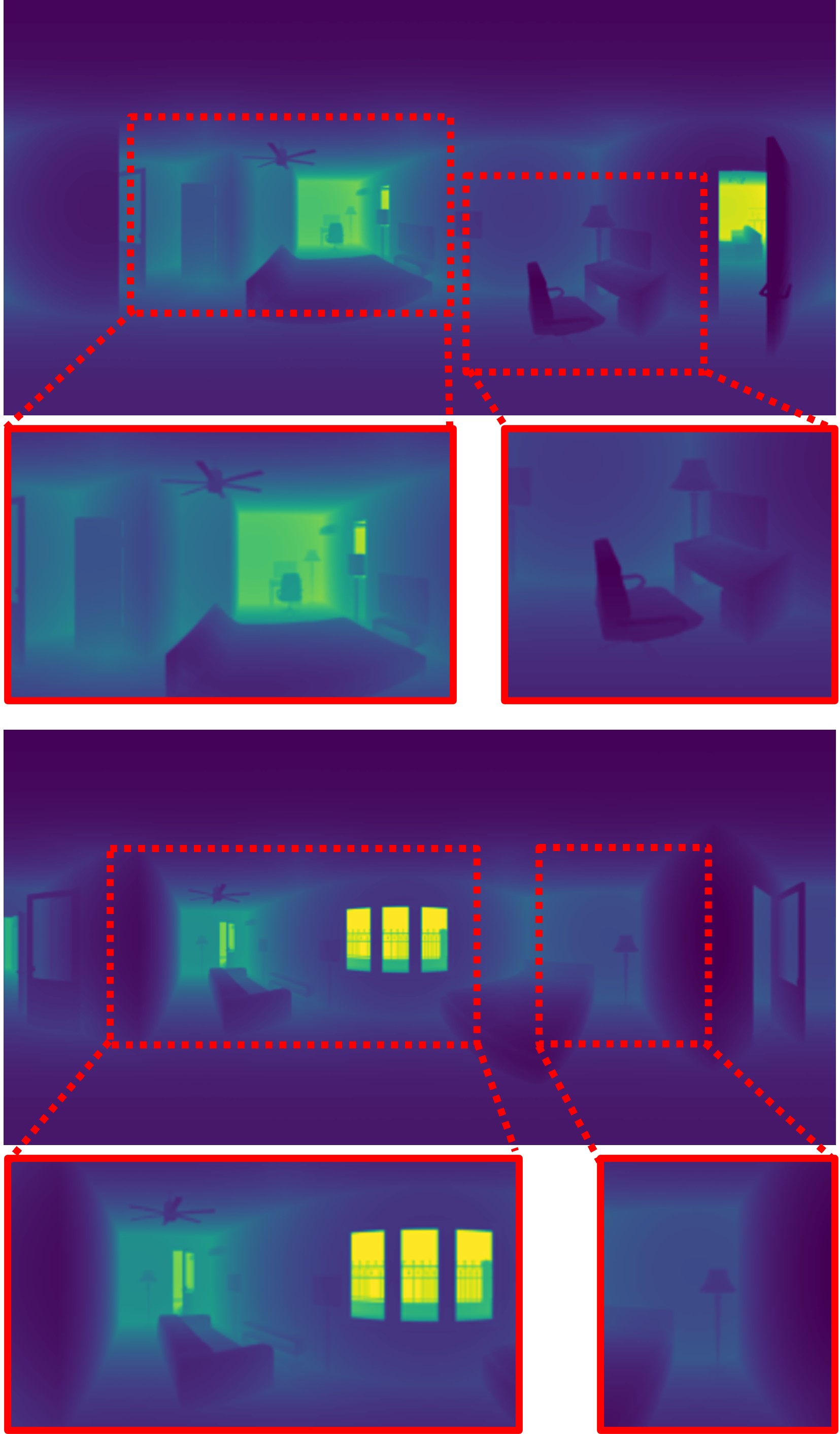}
    \end{subfigure}
    \begin{subfigure}{\piclength}
    \label{fig:suncg:BiFuse} 
    \includegraphics[width=\textwidth]{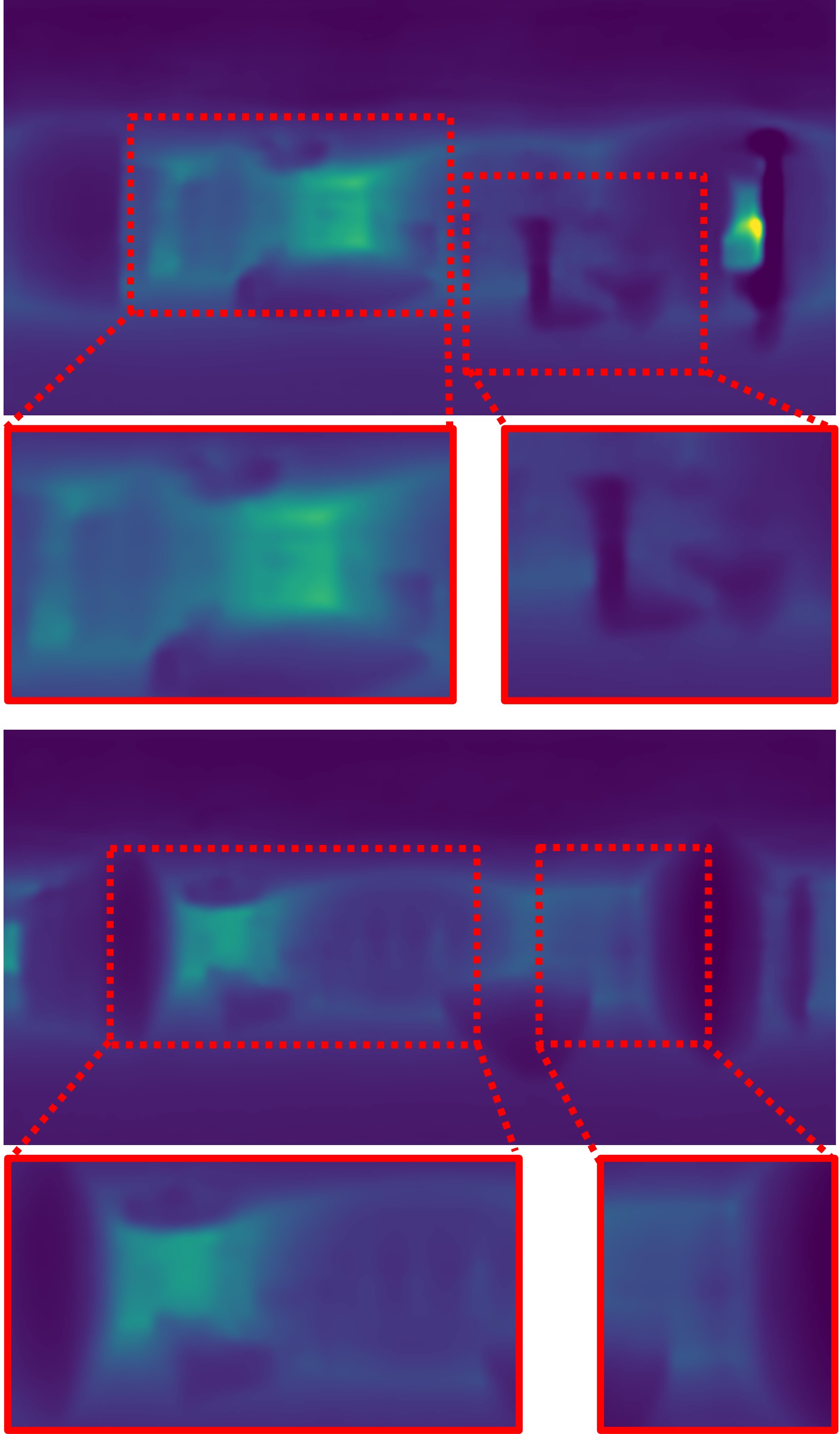}
    \end{subfigure}
    \begin{subfigure}{\piclength}
    \label{fig:suncg:mine}
    \includegraphics[width=\textwidth]{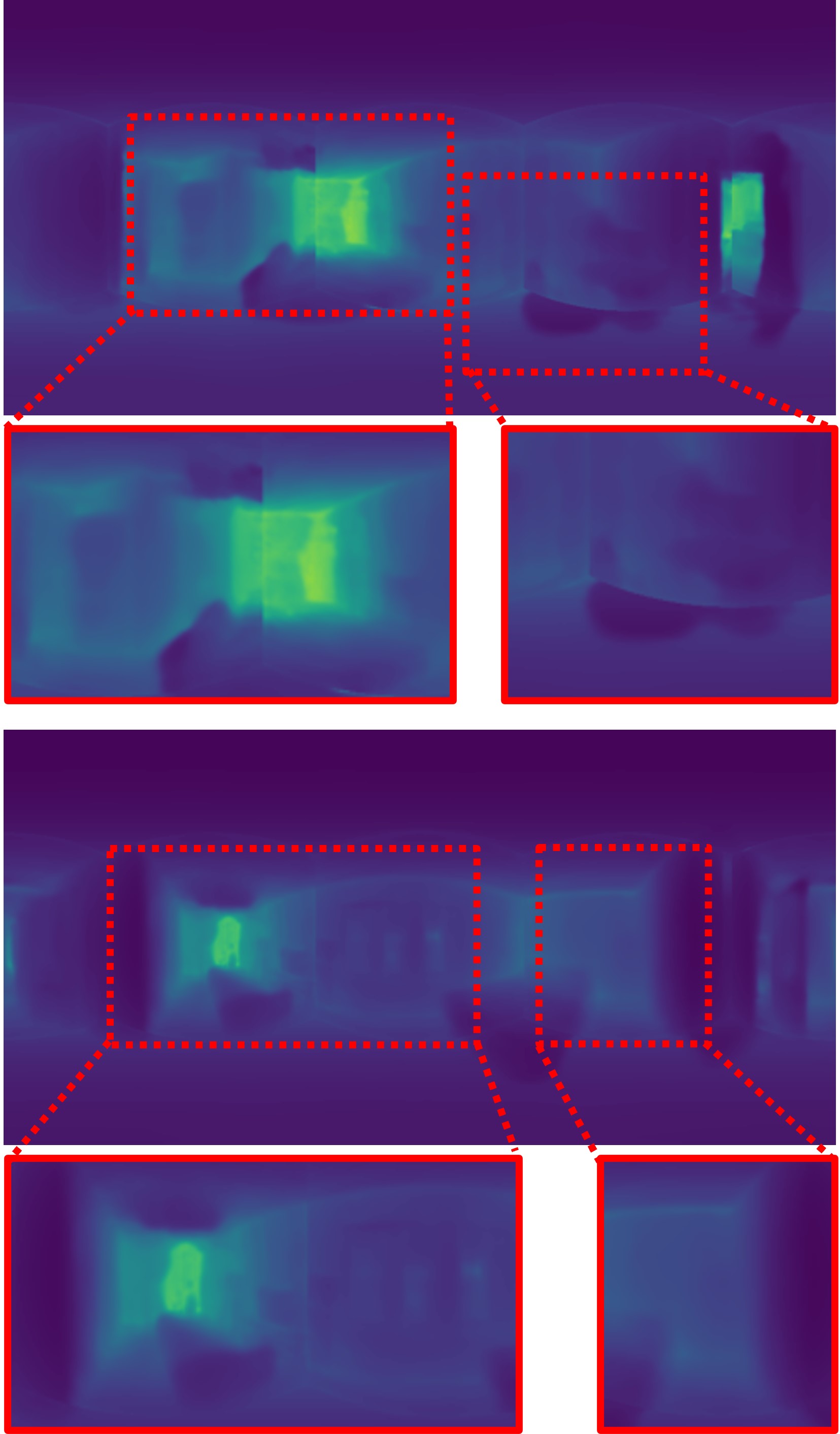}
    \end{subfigure}
    \begin{subfigure}{\piclength}
    \label{fig:suncg:ours}
    \includegraphics[width=\textwidth]{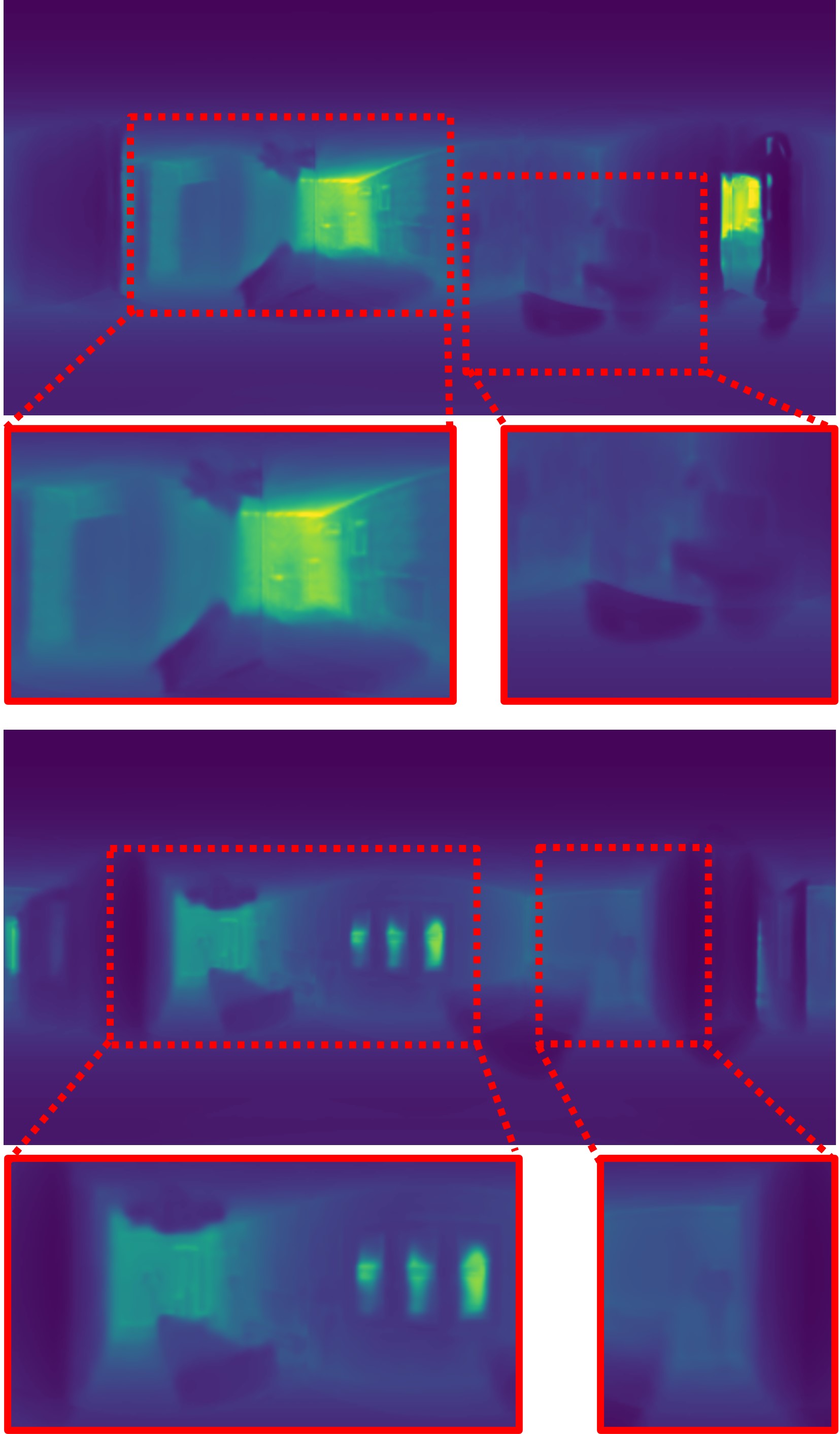}
    \end{subfigure}
    \\
    \vspace{0.1cm}
    \raisebox{0.35in}{\rotatebox{90}{{\scriptsize Matterport3D}}}
    \centering
    \begin{subfigure}{\piclength}
    \label{fig:Matterport3D:rgb} 
    \includegraphics[width=\textwidth]{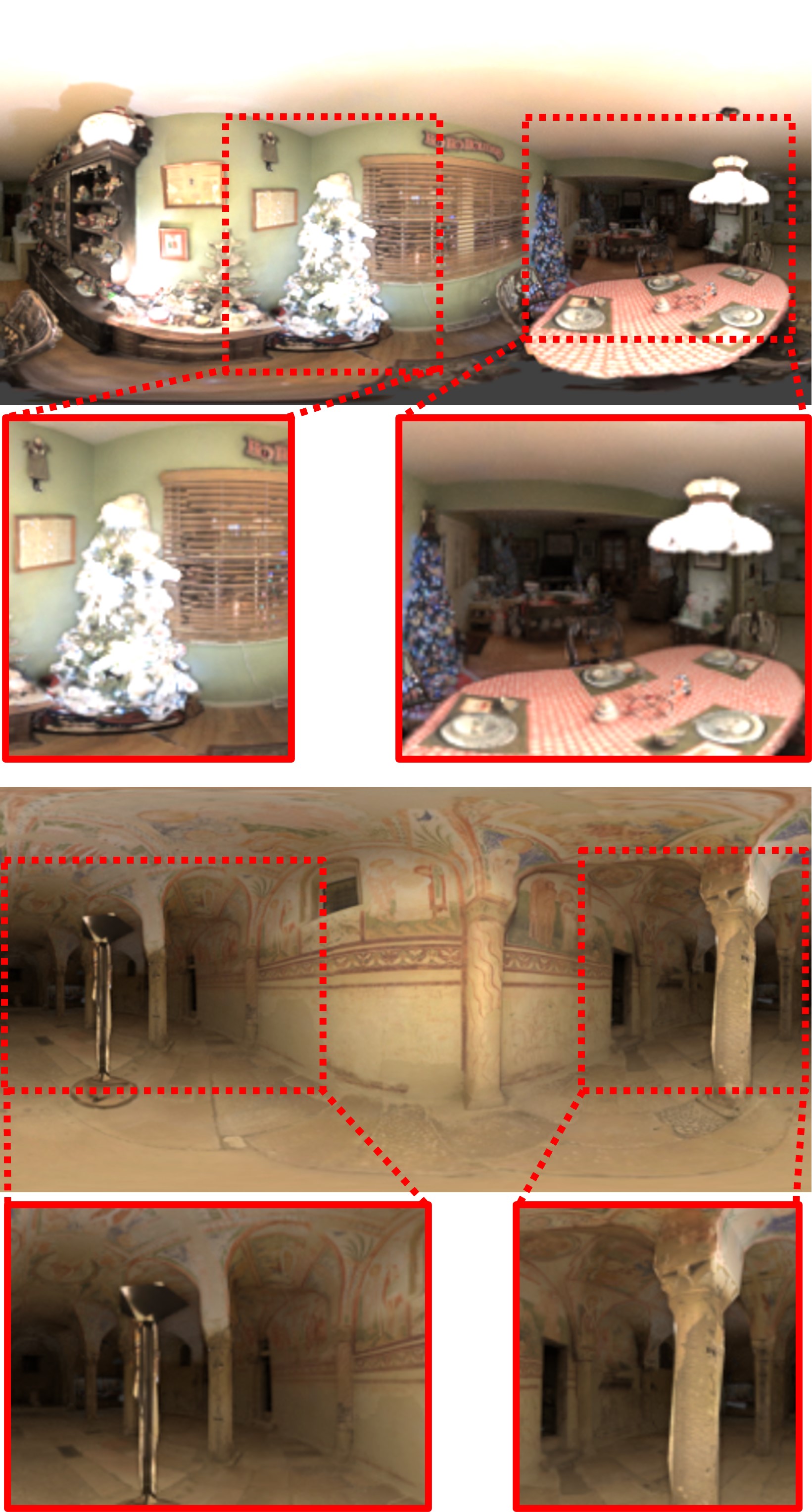}
    \end{subfigure}
    \begin{subfigure}{\piclength}
    \label{fig:Matterport3D:gt}
    \includegraphics[width=\textwidth]{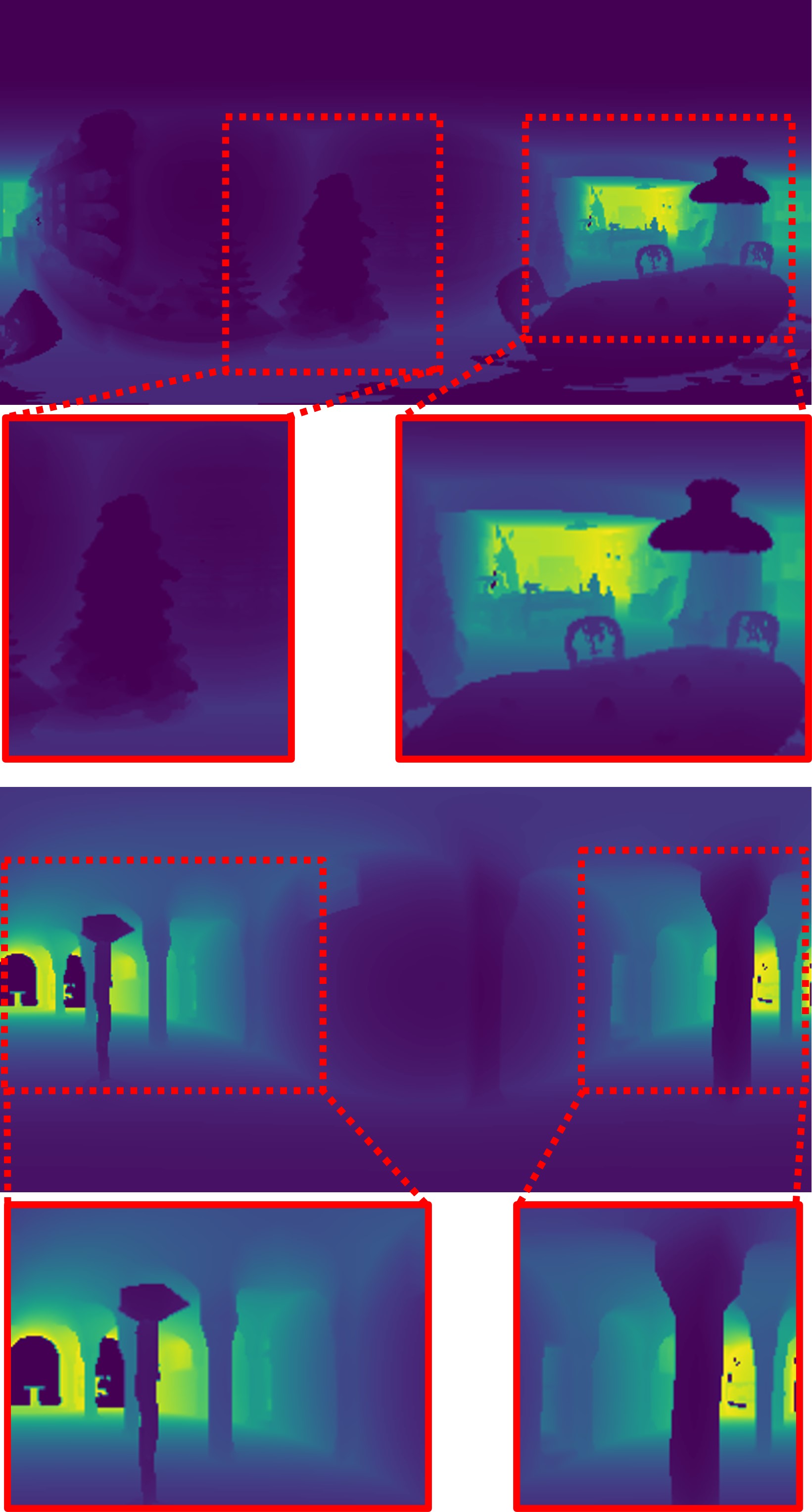}
    \end{subfigure}
    \begin{subfigure}{\piclength}
    \label{fig:Matterport3D:BiFuse} 
    \includegraphics[width=\textwidth]{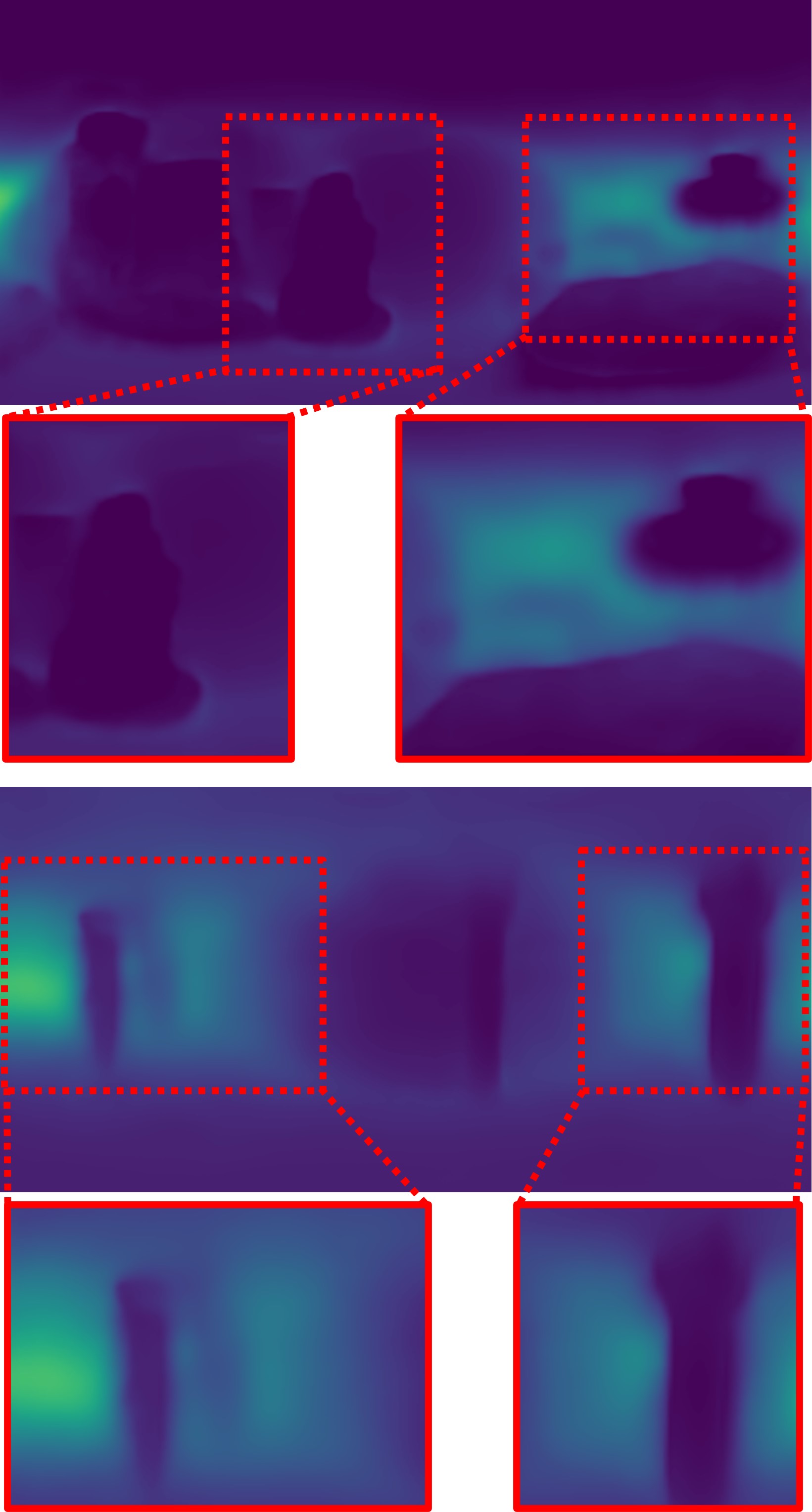}
    \end{subfigure}
    \begin{subfigure}{\piclength}
    \label{fig:Matterport3D:mine}
    \includegraphics[width=\textwidth]{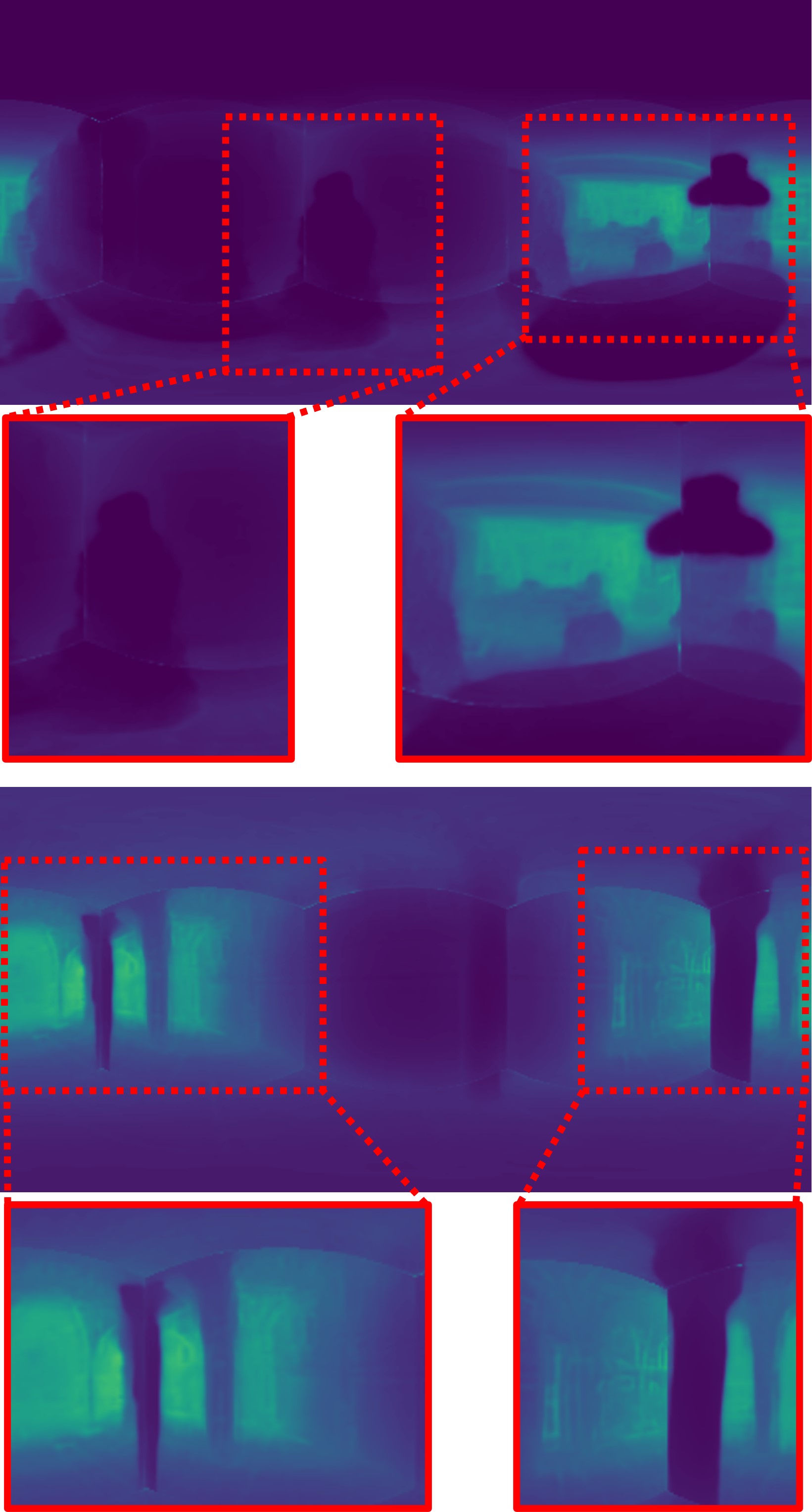}
    \end{subfigure}
    \begin{subfigure}{\piclength}
    \label{fig:Matterport3D:ours}
    \includegraphics[width=\textwidth]{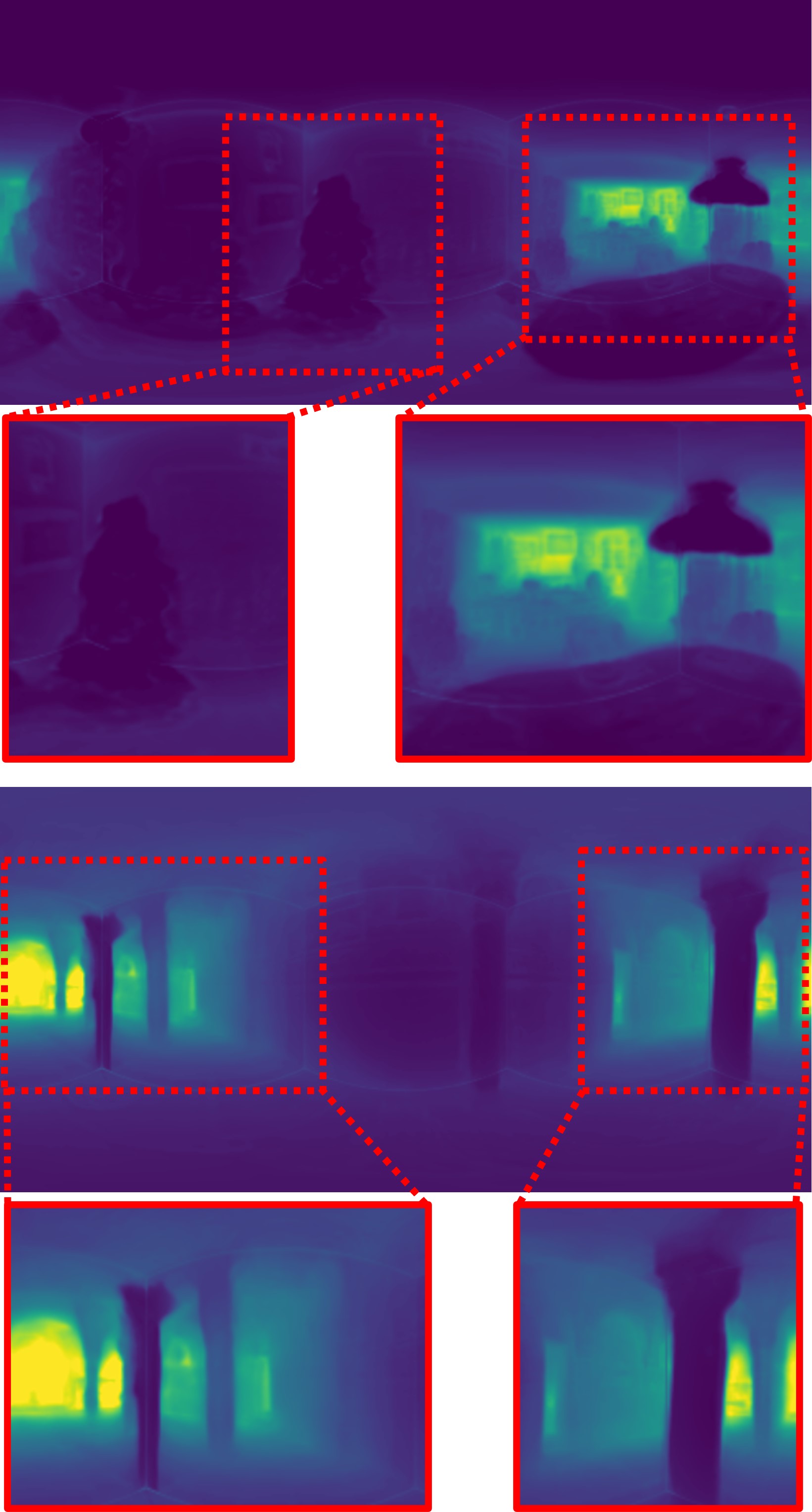}
    \end{subfigure}
    \\
    \vspace{0.1cm}
    \raisebox{0.4in}{\rotatebox{90}{{\scriptsize Stanford2D3D}}}
    \centering
    \begin{subfigure}{\piclength}
    \includegraphics[width=\textwidth]{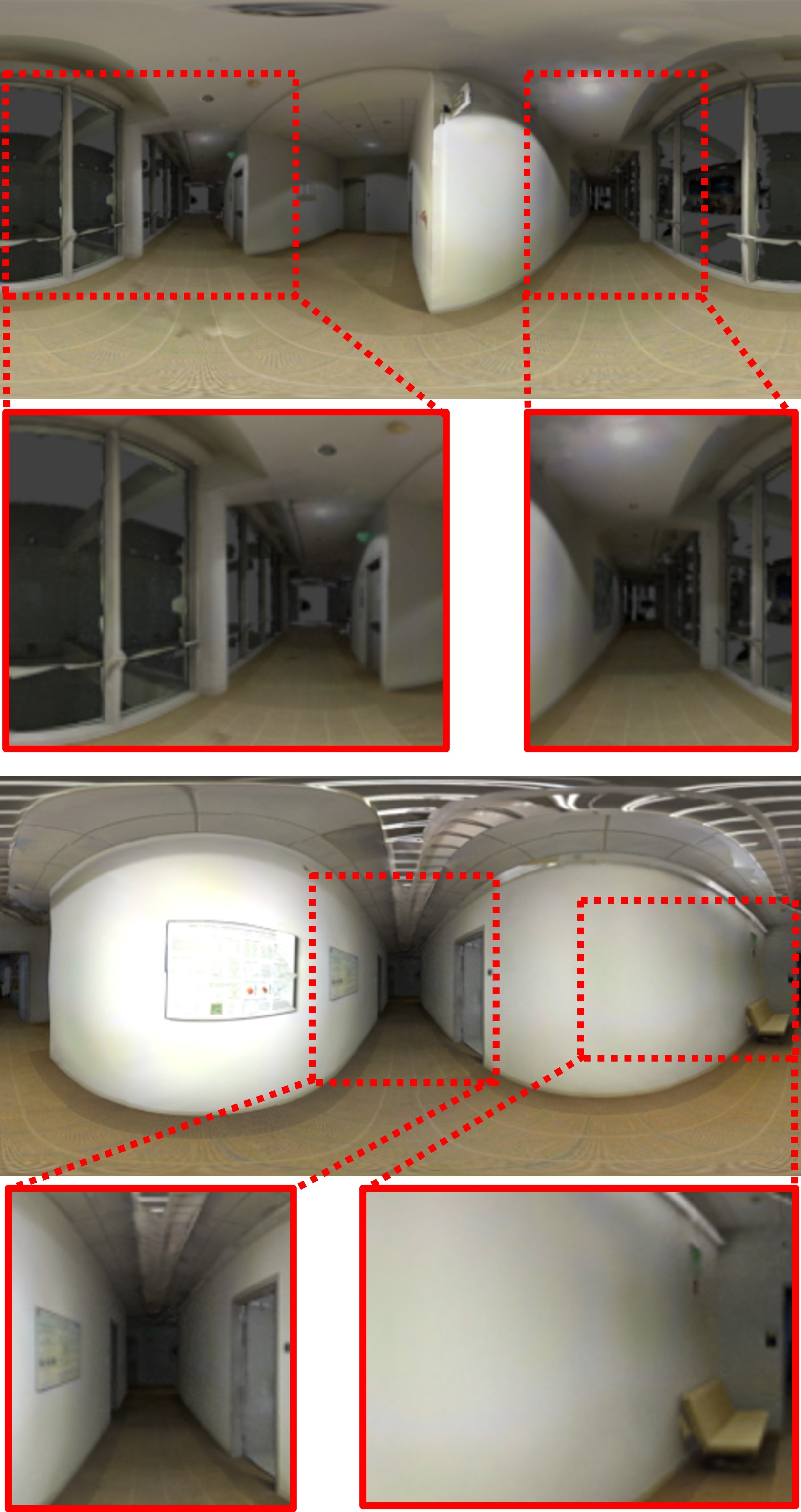}
    \caption{RGB}
    \label{fig:stanford:rgb} 
    \end{subfigure}
    \begin{subfigure}{\piclength}
    \includegraphics[width=\textwidth]{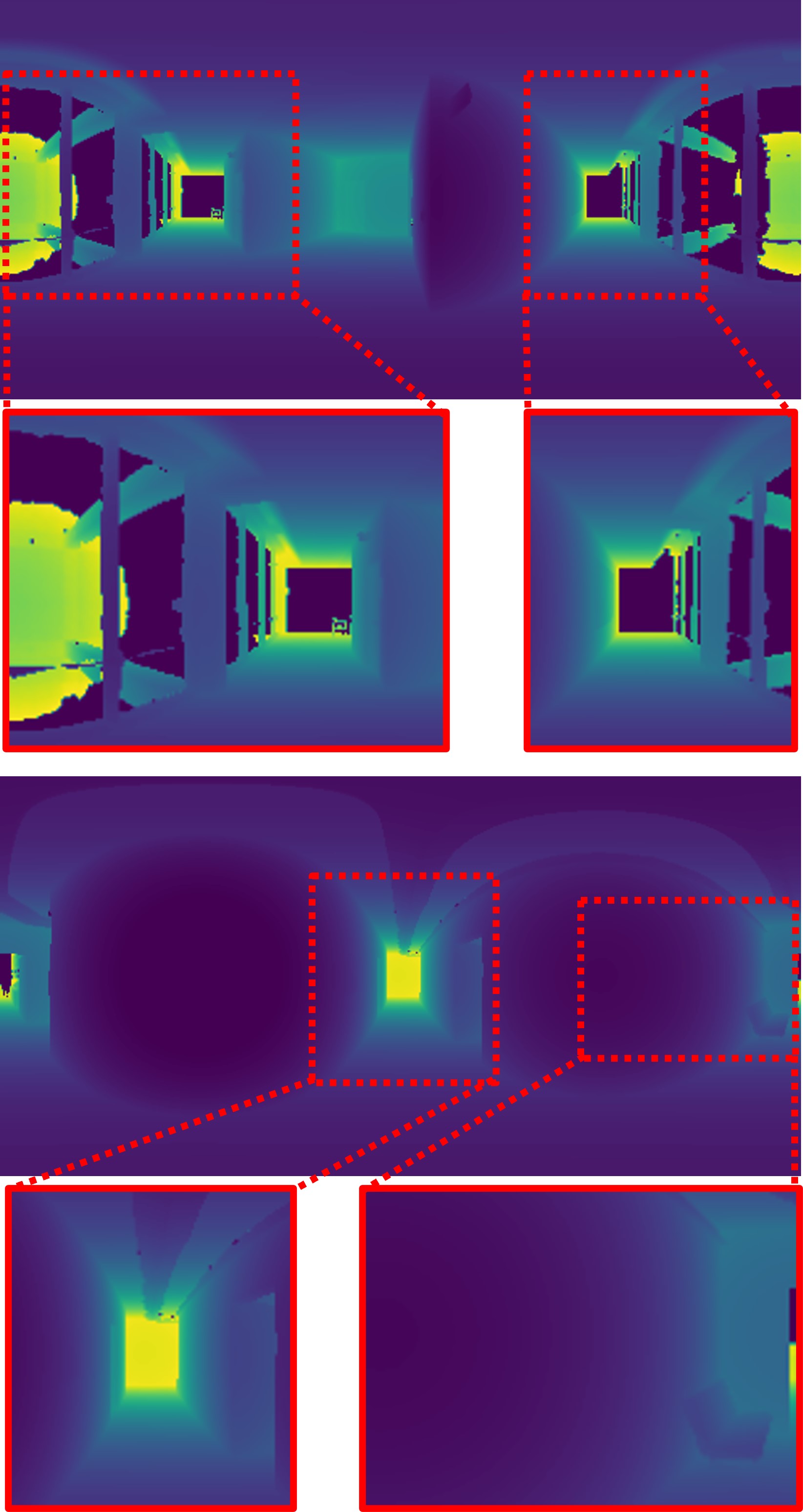}
    \caption{GT}
    \label{fig:stanford:gt}
    \end{subfigure}
    \begin{subfigure}{\piclength}
    \label{fig:stanford:BiFuse} 
    \includegraphics[width=\textwidth]{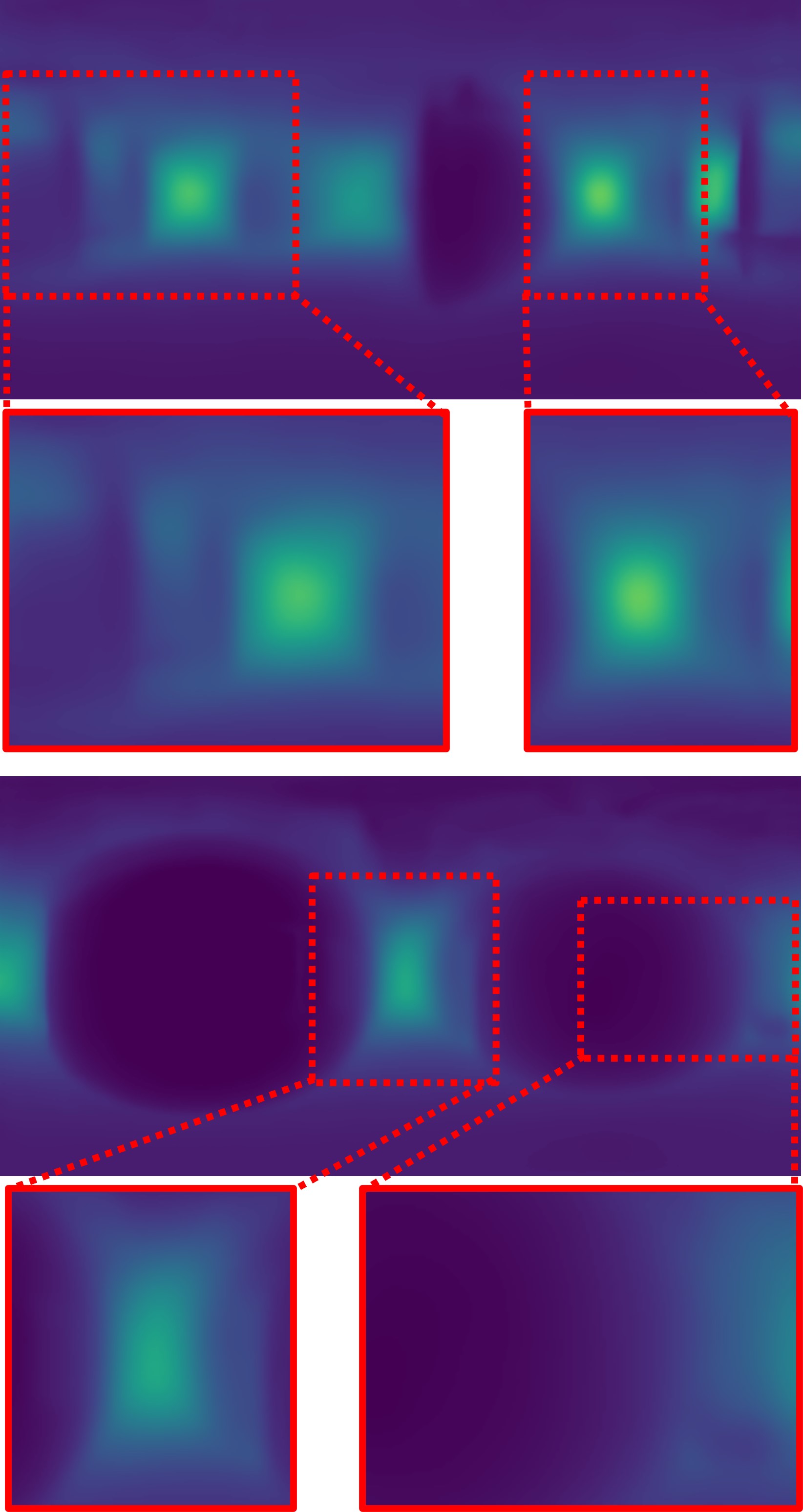}
    \caption{ Bifuse++ \cite{Wang2022BiFuseSA}}
    \end{subfigure}
    \begin{subfigure}{\piclength}
    \includegraphics[width=\textwidth]{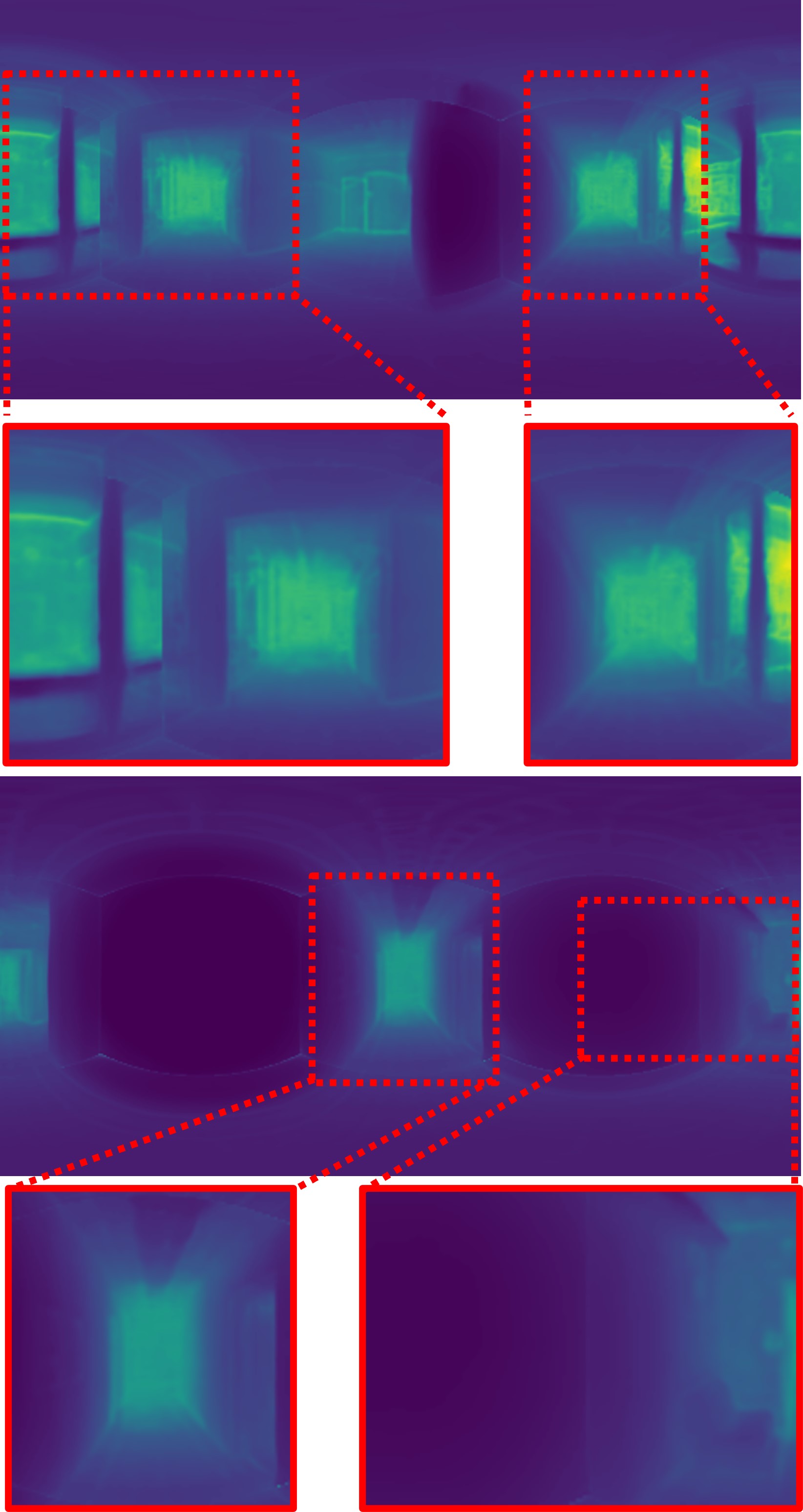}
    \caption{MINE \cite{li2021mine}}
    \label{fig:stanford:mine}
    \end{subfigure}
    \begin{subfigure}{\piclength}
    \includegraphics[width=\textwidth]{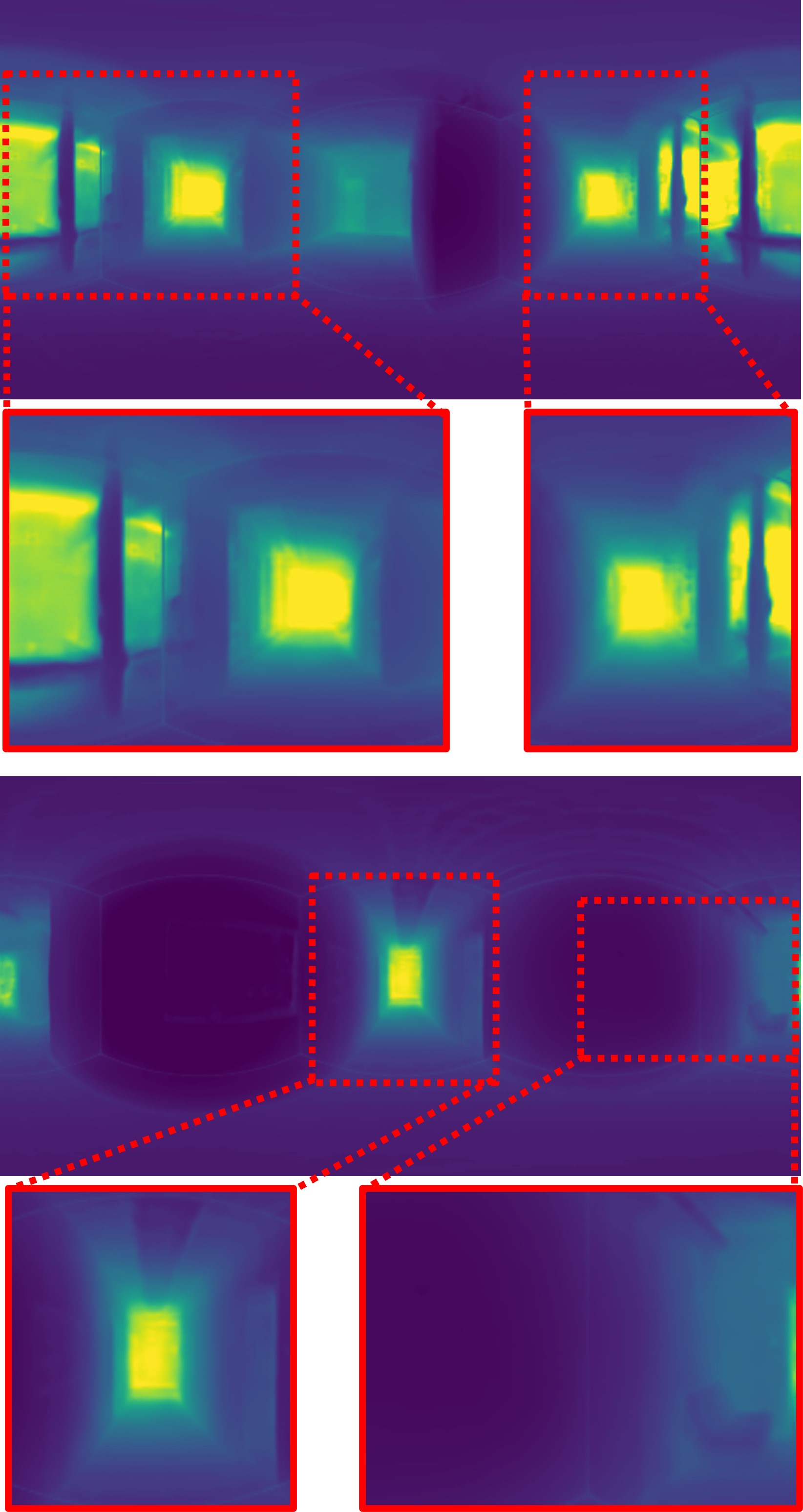}
    \caption{Ours}
    \label{fig:stanford:ours}
    \end{subfigure}
    \\
    \vspace{-8pt}
    \caption{\textbf{Qualitative comparisons between ours, Bifuse++, MINE and SPDET} on the PanoSUNCG, Matterport3D and Standford2D3D datasets, respectively. \subref{fig:stanford:rgb} is for the inputs and \subref{fig:stanford:gt} is for the corresponding ground truth depth maps. \textcolor{red}{(c)-(e)} are for the predictions of compared methods and ours. 
    }
    \label{fig:Matterport3D}
    \vspace{-7pt}
\end{figure*}

\section{Experiments}
\subsection{Datasets}
\noindent{\bf PanoSUNCG}.
PanoSUNCG is a synthetic dataset derived from the SUNCG dataset. It contains a large amount of 360$^\circ$ videos with depth ground truths and camera motion. In detail, it records a total of 25K panoramas. Meanwhile, PanoSUNCG provides realistic and diverse indoor scenes with various camera trajectories and lighting conditions. Following~\cite{Wang2022BiFuseSA}, we choose 80 scenes as the training dataset and 23 as the testing dataset. The resolution is 512 $\times$ 1024.

\noindent{\bf 3D60}. 
The 3D60 \cite{zioulis2019spherical} dataset is different modalities, including RGB panoramas, depth maps, and normal maps. It is generated from existing 3D datasets of indoor scenes, Matterport3D \cite{Matterport3D}, Stanford2D3D \cite{armeni2017joint} and SunCG \cite{song2016ssc}, using ray-tracing techniques. The dataset comprises over 20,000 viewpoints of trinocular stereo pairs, with virtual cameras positioned at the central, right, and up viewpoints, and a fixed baseline distance of 0.26 meters. Following \cite{zhuang2023spdet}, we utilize the real-world subset and follow the official splits.

\subsection{Evaluation}
\noindent{\bf Baselines}. 
To evaluate our method, we conduct comparisons against the state-of-the-art (SOTA) self-supervised methods. BiFuse++~\cite{Wang2022BiFuseSA} is a novel approach for estimating the depth from a single $360^{\circ}$ video. It uses bi-projection fusion, which is a technique that leverages information from equirectangular images and corresponding cubic images to improve the estimation performance. MINE~\cite{li2021mine} introduces a novel method for generating realistic and high-quality novel views from a single image using a continuous depth generalization of the MPIs technique and the NeRF technique. SPDET~\cite{zhuang2023spdet} is a self-supervised 360$^\circ$ depth estimation method from a single RGB image using a transformer network and a spherical geometry-aware feature. 

For the evaluation metrics, we follow~\cite{BiFuse20,zioulis2018omnidepth} to use the standard metrics to evaluate the performance: mean absolute error (MAE), mean relative error (MRE), root mean square error (RMSE) of linear measures and relative accuracy $\delta_1$, $\delta_1$ and $\delta_3$ (the fraction of pixels where the relative error is within a threshold of $1.25$, $1.25^2$ and $1.25^3$). All errors are calculated in meters.
\begin{table}[t]
\centering
\caption{Ablation study on the effects of the number of image planes. Experiments are constructed on PanoSunCG.}
\label{table:mpi}
   \begin{tabular}{ccccc}
\hline\noalign{\smallskip}
Scheme&MAE$\downarrow$ & MRE $\downarrow$ & RMSE $\downarrow$ & $\delta_1 \uparrow$\\
\noalign{\smallskip}
\hline
\noalign{\smallskip} 
 16 &0.2347 &0.1674 &0.4499 &0.8323 \\ 
 32 &0.1673&0.1015&0.3839&0.8970\\
\hline
\end{tabular}
\end{table}

\begin{table}[t]
\centering
   \caption{Experimental results on different sampling strategies. Experiments are constructed on PanoSunCG. P.H.S. denotes the Planar Homography Sampling and C.R.S represents Ray-Cube Sampling.}
   \label{table:sampling}
\begin{tabular}{ccccc}
\hline\noalign{\smallskip}
Strategy&MAE & MRE $\downarrow$ & RMSE $\downarrow$ & $\delta_1 \uparrow$\\
\noalign{\smallskip}
\hline
\noalign{\smallskip} 
 P.H.S. &0.1748 &0.0800 &0.3922 &0.9096\\ 
 C.R.S &0.1876 &0.1069 &0.4099 &0.8861 \\ 
 All &\textbf{0.1673}&\textbf{0.1015}&\textbf{0.3839}&\textbf{0.8970}	\\
\hline
\end{tabular}
\vspace{-5pt}
\end{table}

\noindent{\bf Quantitative Results}.
In Table~\ref{table:panosuncg}, we compare our CUBE360 with the advanced scene representation methods and depth estimation methods on 2 realistic datasets and a synthesized dataset. Compared with BiFuse++, 
which employs a neural network to infer the pose information of different viewpoints, our CUBE360 leverages the ground truth (GT) pose information directly. Therefore, we conduct a retraining process with the GT pose and the official implementation of BiFuse++. For SPDET, we directly use the pre-trained model from the official report for evaluation. Furthermore, to maintain consistency with the measurement indicators of BiFuse++, we also evaluate SPDET under the same criteria as BiFuse++ for a fair comparison. To the best of our knowledge, there is no existing work on learning MPIs from a single panoramic image. Therefore, we propose a baseline that uses MINE to estimate the depth information of six different cubic faces and then stitches them together to form the panoramic depth map. It can be observed that our CUBE360 shows comparable performance against the state-of-the-art method. Especially on the PanoSUNCG dataset, our method achieves favorably better results than the state-of-the-art methods across all the evaluation metrics.

\noindent{\bf Qualitative Results}.
As illustrated in Figure~\ref{fig:Matterport3D}, we present the qualitative results on PanoSunCG and the two subsets, Matterport3D and Stanford2D3D, from 3D60. To ensure fair comparisons and better visualization, all depth maps are shown in the same depth range.
From the comparison between Figure~\ref{fig:stanford:mine} and Figure~\ref{fig:stanford:ours}, we can observe that our CUBE360 reduces the artifacts between the generated depth maps of different cubic faces. This improvement is credited to our bi-projection fusion, which combines the information from equirectangular images and cubic images. From Figure~\ref{fig:stanford:mine} and Figure~\ref{fig:stanford:ours}, we can see that our cubic-based method can achieve comparable visual quality to BiFuse++ \cite{Wang2022BiFuseSA}, which is based on predicting depth maps from equirectangular images. 


\subsection{Ablation Study and Analysis}

\noindent{\bf Number of Image Planes}.
Table \ref{table:mpi} reports the effect of the number of image planes in MPIs. The performance of depth prediction is greatly improved when the number of planes increases from 16 to 32. Therefore, we chose 32 planes as the default setting.

\noindent{\bf Sampling Strategy}.
To validate the effectiveness of the proposed sampling strategy, we deploy different renderings for photometric loss construction as the supervision. Results are shown in Table \ref{table:sampling}, where P.H.S. denotes the Planar Homography Sampling, and C.R.S. represents the Ray-Cube Sampling. We can see that C.R.S. outperforms P.H.S. due to the superior ability of cubic-level rendering to capture the representation of the entire scene. The addition of planar-level rendering results yields the best performance, which validates the effectiveness of our dual sampling strategy.

\begin{table}
    \centering
   \caption{Ablation study on components of the proposed blending modules. Experiments are constructed on PanoSunCG.}
 \label{table:gru}
\begin{tabular}{ccccc}
    \hline\noalign{\smallskip}
    Scheme&MAE & MRE $\downarrow$ & RMSE $\downarrow$ & $\delta_1 \uparrow$\\
    \noalign{\smallskip}
    \hline
    \noalign{\smallskip} 
     None&0.1819&0.1107&0.4150 &0.8777 \\
     (1) &0.1758&0.1092&0.3956&0.8857\\ 
     (2) &0.1799 &0.1117 &0.4005 &0.8813 \\ 
     All &0.1673&0.1015&0.3839&0.8970\\
    \hline
    \end{tabular}
    \vspace{-5pt}
\end{table}




\noindent{\bf Compenents of Blending Modules}.
Table~\ref{table:gru} presents the results of our ablation study on the components of the proposed blending modules, evaluated on the PanoSunCG dataset. The "None" row represents the baseline performance without any blending modules. Operation (1), described in Fig.~\ref{fig:padding}, corresponds to the padding blending approach, while operation (2), introduced in Fig.~\ref{fig:blending}, represents Inter-Faces Blending and Cubic-ERP Blending. The "All" scheme combines both operations. The results show that incorporating the blending modules significantly improves performance.

\noindent{\bf Efficiency of the Cubic Field}. Since there is no MSI-based method specifically designed for single panoramas \cite{habtegebrial2022somsi,attal2020matryodshka}, we used a combination of MSI representation and the MINE pipeline (MSI+MINE) as a baseline for comparison. For processing a 512×1024 panorama, our cubic field representation requires 16,834M of video memory during training, which is 1.81 times more memory-efficient than the 30,552M required by MSI+MINE. During inference, our method uses only 4,616M of memory, saving 3.42 times more memory compared to the 15,814M required by MSI+MINE. Additionally, our model achieves better accuracy with lower MAE (0.167 vs. 0.183) and RMSE (0.384 vs. 0.425). These results demonstrate that the cubic field not only significantly reduces memory consumption but also improves accuracy.

\begin{figure*}[t]
    \centering
    \includegraphics[width=0.95\linewidth]{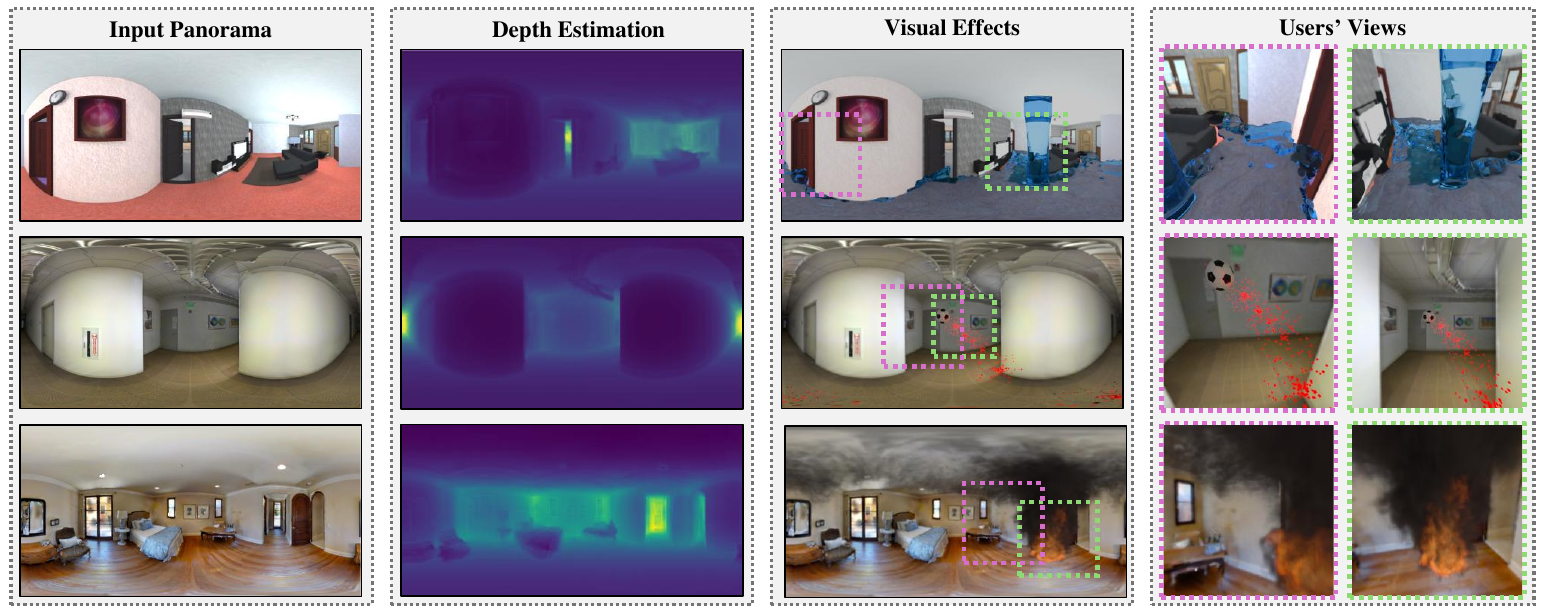}
    \captionof{figure}{Visualization of the integration of visual effects into panoramic images, utilizing 3D geometry information from CUBE360 to enhance scene interaction. The first row demonstrates water flow dynamics, the second row features a bouncing soccer ball, and the third row showcases complex fire and smoke interactions with the environment. Users' views represents the perspective seen through a VR device.}
    \label{fig:ar}
\end{figure*}
\section{Practical Applications}
\begin{figure}[t]
    \centering
    \includegraphics[width=\linewidth]{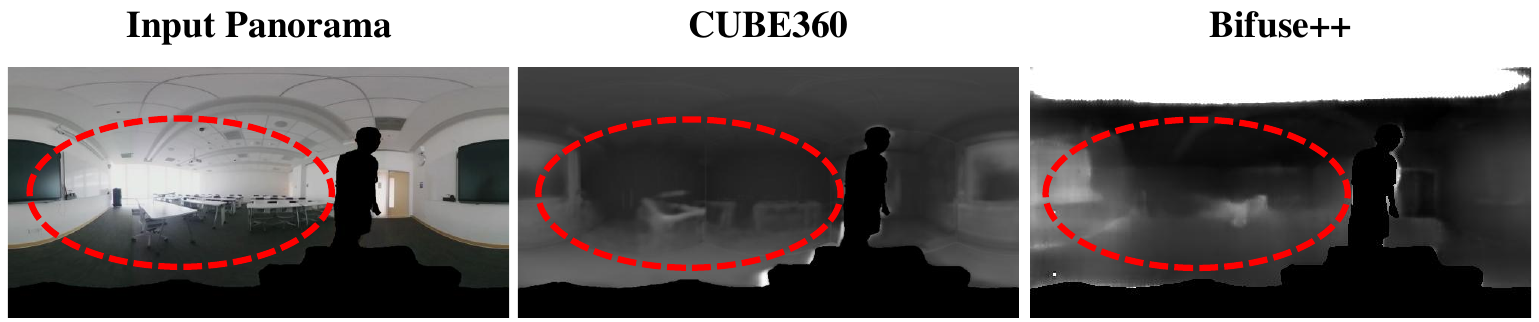}
    \caption{Comparison of depth inference results on real-world panoramic images captured with RICOH THETA Z1 camera. The input panorama (left) is processed using the proposed CUBE360 method (middle), demonstrating enhanced depth prediction quality, especially in challenging regions (highlighted in red), compared to traditional supervised methods (right).}
    \label{fig:real}
\end{figure}

\noindent{\bf Real-World Adaptation and Evaluation}. 
CUBE360 is built on self-supervised learning, allowing it to be fine-tuned on videos captured by consumer-level panoramic cameras without the need for ground truth depth information. This flexibility enhances its practicality and efficiency in real-world applications. We conducted experiments on self-captured real-world videos and compared the results to those produced by supervised methods. As shown in Fig. \ref{fig:real}, our approach outperforms supervised methods, particularly in challenging regions, further demonstrating its effectiveness in practical scenarios. These findings highlight the potential of our method as a more robust and efficient solution for depth inference from consumer-grade panoramic cameras.

\begin{figure}[t]
    \centering
    \includegraphics[width=\linewidth]{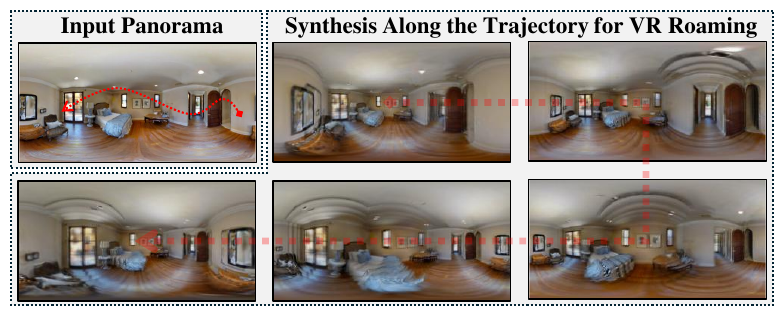}
    \caption{Input panorama and rendered novel views along a predefined trajectory. The top-left shows the original input panorama, while the remaining images present novel views generated along the indicated path (red dotted lines). These novel views highlight the effectiveness of our method in enabling immersive VR experiences with realistic scene exploration.}
    \label{fig:roam}
\end{figure}

\noindent{\bf VR Wondering}. 
CUBE360 constructs a cubic field from a single panorama, capturing both the scene's texture information and geometry. By leveraging this cubic representation, panoramic images from various viewpoints are generated via neural rendering techniques. This approach allows for seamless scene navigation, enabling users to virtually "roam" through the reconstructed environment. As shown in Fig. \ref{fig:roam}, novel views are rendered along a predefined trajectory, showcasing the scene from different positions. Our method achieves a real-time rendering speed of 414 fps on an NVIDIA RTX A6000, which comfortably exceeds the frame rate requirements of mainstream display devices. Full video results are included in the supplementary material for further reference.

\noindent{\bf Visual Effects}. CUBE360 extracts a high-quality depth map from a single panoramic image, which is then converted into a 3D mesh that captures the global structure of the entire scene. This detailed 3D geometry serves as the foundation for integrating visual effects, such as simulating water flow across surfaces, the interaction of a bouncing soccer ball with the environment, and the dynamic behavior of fire and smoke as they respond to the scene’s structure. Fig. \ref{fig:ar} illustrates these effects, including user views seen through VR devices. Full video demonstrations are available in the supplementary.

\section{Conclusion and Future Work}
In this paper, we proposed a novel method, CUBE360, for the 360 depth estimation from a single panorama in a self-supervised manner. Our method learns a cubic field representation to model the color and density information of a holistic scene. We introduced a novel sampling strategy that enables novel view synthesis at both cubic and planar scales from the cubic field. Besides, we proposed a attention-based blending model that integrates cross-face features to generate the cubic field. The proposed CUBE360 showed its superiority over SOTA methods in terms of accuracy and generalization capacity On both synthetic and real-world datasets. Furthermore, the demonstrated practical applications highlight the utility of the proposed method.



\bibliographystyle{abbrv-doi}

\bibliography{template}
\end{document}